\documentclass[journal]{IEEEtran}
\pdfoutput=1
\usepackage{cite}
\usepackage{amsmath,amssymb}
\usepackage{booktabs}
\usepackage{array}
\usepackage{tabularx}
\usepackage{url}
\usepackage{tikz}
\usepackage{pgfplots}
\pgfplotsset{compat=1.18}
\usepgfplotslibrary{groupplots}
\usetikzlibrary{positioning,arrows.meta}
\usepackage[hidelinks]{hyperref}
\hypersetup{pdftitle={Unilateral Adoption of LLM-Written Policy Weights in a Multi-UAV Aerial Base Station Game}, pdfauthor={Yu-Chen Hsu and Li-Hsing Yen}}
\definecolor{cBlue}{RGB}{0,114,178}
\definecolor{cOrange}{RGB}{230,159,0}
\definecolor{cGreen}{RGB}{0,158,115}
\definecolor{cRed}{RGB}{213,94,0}
\definecolor{cPurple}{RGB}{204,121,167}
\pgfplotsset{
  paper/.style={
    width=\columnwidth, height=0.62\columnwidth,
    tick label style={font=\footnotesize}, label style={font=\footnotesize},
    title style={font=\footnotesize, yshift=-1ex},
    legend style={font=\footnotesize, draw=none, fill=white, fill opacity=0.85, text opacity=1, inner sep=1pt},
    legend cell align=left,
    grid=major, grid style={gray!22},
    every axis plot/.append style={line width=0.8pt},
    tick align=outside, tick pos=left,
  },
}

\begin{document}
\bstctlcite{IEEEexample:BSTcontrol}

\title{Unilateral Adoption of LLM-Written Policy Weights in a Multi-UAV Aerial Base Station Game}

\author{Yu-Chen~Hsu and Li-Hsing~Yen%
\thanks{The authors are with the Institute of Computer Science and Engineering, National Yang Ming Chiao Tung University, Hsinchu, Taiwan.}}

\maketitle

\begin{abstract}
Unmanned aerial vehicles (UAVs) serving as aerial base stations share users, channels and charging stations, so one UAV's gain can be another's loss. We study a noncooperative game in which UAVs of different operators change policies only for their own benefit. Each UAV earns a per-step reward that mainly counts the signal-to-interference-plus-noise ratio of its users. A crash incurs a penalty and ends future rewards. A UAV's score is the episode mean. In each attempt, a large language model (LLM) that can read but not run the simulator writes a program setting every weight of one UAV's policy network. This UAV adopts the LLM-written policy weights only if its own score rises by at least 1\% while the others keep theirs, a rule we call unilateral adoption. An attempt costs one LLM answer and 10 simulated episodes, and adopted weights run without LLM calls. In a 5-UAV run with static users, unilateral adoption stopped after 70 attempts and raised the fleet-mean score by 25\%, from 84.7 to 105.9, although one UAV ended below its starting score. A 30-UAV run imposing each adopted weight set on all UAVs cycled below its starting fleet mean. Both runs started from one shared LLM-written weight set. Without an energy guard that forces low-battery UAVs to charge, these weights kept all 30 UAVs alive on two training maps and a held-out map. Five multi-agent reinforcement learning baselines lost 22.6 to 30 of 30 UAVs per episode on the training maps. We ran each configuration once.
\end{abstract}


\section{Introduction}
\label{sec:intro}

\IEEEPARstart{U}{nmanned} aerial vehicles (UAVs) that carry small base stations can provide wireless coverage where terrestrial infrastructure is damaged or absent~\cite{Mozaffari2019}. Key challenges of such aerial base stations include three-dimensional deployment, trajectory optimization, resource allocation, interference management and energy efficiency~\cite{Mozaffari2019}. When several UAVs serve the same area, energy and interference couple their decisions. A UAV whose battery runs out crashes and stops serving its users. Each UAV must trade service time against time spent at charging stations, whose slots the UAVs share. Ground users associate with UAVs by their signal-to-interference-plus-noise ratio (SINR). The movement and transmission of one UAV thus change the interference and the users of its neighbors.

The UAVs over one disaster area may belong to different operators. Each operator then cares only about the service of its own UAV, while the UAVs compete for users, channels and charging slots. A control change that helps one UAV can take users or charging slots from the others. The control of the UAVs then becomes a noncooperative game, and it is not clear in advance where a sequence of such changes leads.

Many studies address the deployment, trajectories and resources of multi-UAV aerial base stations. Joint optimization methods choose trajectories, user association and transmit power under energy limits to maximize the total rate~\cite{Wang2025}. Game-theoretic studies model UAV positioning and user association as noncooperative or potential games~\cite{ElHammouti2020, ChouYen}. Multi-agent reinforcement learning (RL) trains policies for cooperative tasks~\cite{MAPPO}, including UAVs that share charging stations~\cite{YangThesis}. Game-theoretic RL computes best responses against populations of opponents~\cite{PSRO}. Chou and Yen~\cite{ChouYen} formulated the placement of UAV base stations as a game in which each UAV chooses its next position, and its utility depends on the spectral efficiency of its users. They did not model battery depletion or recharging. Yang~\cite{YangThesis} studied 30 energy-limited UAVs that share three charging stations with limited slots. That work trained one shared policy for a common objective with multi-agent proximal policy optimization (MAPPO)~\cite{MAPPO}.

Large language models (LLMs) have also been used as iterative optimizers~\cite{OPRO, Eureka, AlphaEvolve}. Recent reports show that LLMs can write or edit the weights of a neural network without gradient descent~\cite{Takehi2026, WeightLoop, Gross2026, Zhang2026, Hu2026}. Takehi et al.~\cite{Takehi2026} call such writing \emph{weight-space programming}. Other work lets LLMs write game strategies as programs from evaluation feedback~\cite{CSRO, Gallego2026}. Takehi et al.\ let an LLM write Python code that builds the weights of a Go network, and they kept a change only if the network won more games against fixed opponents. Zhang et al.~\cite{Zhang2026} and Hu et al.~\cite{Hu2026} applied weight editing within an evaluation budget to robot policies and language models. In games among LLM agents, Opponent-Aware Selective Evolution~\cite{OASE2026} keeps a revision of the text skills of an agent only if its estimated payoff gain meets a threshold. Lin et al.~\cite{LinPIBR2025} let an LLM refine the policy code of two agents in turn in cooperative tasks. Tang et al.~\cite{Tang2026} and Nguyen et al.~\cite{Nguyen2026} add an LLM to UAV games. In both works the LLM sets utility weights or resource prices of the game and does not write the weights of a policy network.

However, these works do not cover UAVs whose policies are chosen by self-interested operators. The UAV games of Chou and Yen and of El Hammouti et al.\ choose positions or associations and do not model repeated charging or crashes~\cite{ChouYen, ElHammouti2020}. The LLM-assisted UAV games set game parameters rather than policy weights~\cite{Tang2026, Nguyen2026}. The study with repeated charging uses one shared cooperative policy~\cite{YangThesis}. If each UAV instead chooses a complete policy, an RL best response can require millions of game simulations~\cite{CSRO}. It has to be trained again whenever another player changes its policy~\cite{PSRO}. On the LLM side, the work on LLM-written weights improves a single network against a fixed objective or fixed opponents~\cite{Takehi2026, Gross2026, Zhang2026, Hu2026}. The LLM game frameworks write strategy code rather than network weights~\cite{CSRO, Gallego2026, LinPIBR2025}, adopt text strategies of LLM agents~\cite{OASE2026} or study cooperative tasks without a payoff test~\cite{LinPIBR2025}. We found no work in which several self-interested players adopt LLM-written policy weights in turn, each for its own payoff. It remains unknown whether such a process stops and how the gain of one UAV affects the other UAVs and the fleet. An \emph{energy guard} is a rule that forces a UAV with a low battery back to a station. It is also unknown whether an LLM that may read but not run the code of a simulator can write weights under which no UAV crashes without an energy guard.

This paper studies the unilateral adoption of LLM-written policy weights in a game among UAV aerial base stations. Each UAV earns a reward in every step, and its score is the mean of this reward over an episode. The strategy of each UAV is the complete weight set of its policy network, and its payoff is its own score (Section~\ref{sec:problem}). The fleet-mean score is the mean of the scores of all UAVs. In each attempt, an LLM writes a program that sets every weight of the policy of one UAV. This UAV adopts the new weights only if its own score rises by at least 1\% while the other UAVs keep theirs (Section~\ref{sec:adoption}). A gain against a fixed fleet may not last when all UAVs change in the same way. To test whether such a gain lasts, we compare unilateral adoption with a procedure that imposes each adopted weight set on all UAVs. To obtain the starting weights of the game, we let the LLM write one weight set for the whole fleet, in a single answer and in a feedback loop. We compare these weights with MAPPO baselines. The LLM may read the simulator code but may not run it, because a player in a game is assumed to reason about the others rather than try out every strategy. The paper asks three questions.
\begin{itemize}
\item Does unilateral adoption stop, and how does an adoption affect the other UAVs and the fleet-mean score? How does the outcome differ when each adopted weight set is imposed on all UAVs?
\item Can an LLM that can read but not run the simulator write weights under which no UAV crashes without an energy guard?
\item How does feedback from earlier rounds change the fleet-mean score compared with a single answer?
\end{itemize}

To our knowledge, this paper is the first to study several self-interested players that adopt LLM-written policy weights in turn. The approach also differs from RL in cost. An attempt costs one LLM answer and 10 simulated episodes rather than a trained best response. The adopted weights run as a small network without LLM calls. In a 5-UAV run with static users, unilateral adoption stopped after 70 attempts with 23 adoptions and raised the fleet-mean score by 25\%, from 84.7 to 105.9. The first adoptions lowered the fleet mean, and one UAV ended below its starting score. A 30-UAV run that imposed each adopted weight set on all UAVs did not stop by itself. It cycled through three programs, and its fleet-mean score stayed between 131.4 and 133.4, below its starting value of 149.1 (Section~\ref{sec:games}).

We evaluate the weights on three worlds, which are layouts of charging stations and user hotspots. Two of them are training worlds, and the third is a held-out world that no adoption test or round selection used. Without an energy guard, all four weight sets that the LLM wrote for the fleet-mean score of 30 UAVs, among them the starting weights of the game, kept all UAVs alive on all three worlds. In the feedback loop, round 1 had no feedback yet and scored 132.9. Round 13 scored 149.1, which is 12\% higher. A separate single answer under the same task condition scored 137.9 in its own run's evaluation. Without the guard, all five MAPPO baselines lost 22.6 to 30 of their 30 UAVs per episode on the training worlds. The LLM, however, could read the energy and charging logic of the simulator (Sections~\ref{sec:fleet} to~\ref{sec:rounds}). Every configuration was run once, and Section~\ref{sec:threats} states what these single runs do and do not establish.

Section~\ref{sec:related} reviews related work. Section~\ref{sec:system} presents the system model, and Section~\ref{sec:problem} formulates the problem. Section~\ref{sec:approach} describes the LLM weight writing and the game loops. Section~\ref{sec:setup} gives the setup, Section~\ref{sec:experiments} reports the results, Section~\ref{sec:threats} discusses them, and Section~\ref{sec:conclusion} concludes the paper.

\section{Related Work}
\label{sec:related}

\subsection{UAV Aerial Base Stations}

Chou and Yen~\cite{ChouYen} formulate the placement of UAV access points as a game. The strategy of a UAV is its next position, and its utility is the number of its associated terminals multiplied by their average spectral efficiency. The sum of these utilities is the social welfare of their game and is proportional to the average spectral efficiency that they evaluate. The radio model, the association rule of the users and the service component of the reward of this paper are adapted from that work, with the changes that Section~\ref{sec:radio} lists. The service component of a UAV here has the form of the utility of a UAV there. It uses the SINR of Eq.~\eqref{eq:sinr}, which adds noise and counts as interferers only the transmitting UAVs on the same channel, and it caps the SINR. Yang~\cite{YangThesis} trains one shared MAPPO policy for energy-limited UAVs that share three charging stations of limited capacity, and this paper uses his area, stations, charging rules and energy constants. The movement and flight-energy calculations of Section~\ref{sec:system} are ours, and the charging calculation of Appendix~\ref{appx:sim} is that of Yang's code. El Hammouti et al.~\cite{ElHammouti2020} formulate joint UAV positioning and user association as a potential game in which the utility of a UAV is its marginal contribution to the total rate. Wang et al.~\cite{Wang2025} jointly optimize user association, trajectories and transmit power to maximize the total communication rate of the system under a sensing constraint. Of these four works, only Yang models repeated charging at stations with limited slots. Wang et al.\ include flight and transmit energy and a battery limit, Chou and Yen count the traveling distance as the cost of a deployment, and El Hammouti et al.\ do not model energy. A separate utility for each UAV is therefore not new in this paper. What differs is the decision of a player and how it is evaluated. A player here chooses the complete weight set of a policy, and its payoff is the score of that policy over whole episodes with energy use, charging and crashes.

Two recent works add an LLM to a game among UAVs. Tang et al.~\cite{Tang2026} use two exact potential games. In the first, each UAV chooses its links to other UAVs, and its utility counts its links, its interference and its communication energy. In the second, each UAV chooses its position, its transmit power and its ground users, and its utility combines its throughput, energy and delay. The potential function of each game is the sum of the utilities, and the authors state that the change in the utility of a UAV that alone changes its strategy equals the change in the potential function. Their stated aim is network-level performance. In an initialization phase, an LLM generates the utility functions and their weight coefficients, and a network builder can give feedback, after which the LLM regenerates the weight coefficients until the builder stops. Nguyen et al.~\cite{Nguyen2026} model delivery and monitoring UAVs that draw resources from ground base stations as a noncooperative game, in which aircraft of the same service cooperate and aircraft of different services compete. An LLM adjusts the utility weights and the resource prices of this game to the service requirements. Pretrained multi-agent reinforcement-learning policies take the new parameters as input, so they are not retrained within the range that they support. In both works the LLM sets parameters of the game and does not write the weights of a policy network.

\subsection{LLMs That Write or Edit Network Weights}

The term \emph{weight-space programming} comes from a blog post by Takehi et al.~\cite{Takehi2026}, which reports that a frontier LLM can read and write the weights of a neural network without gradient descent. In its one-shot experiments the LLM repairs a control network of 171,664 weights by changing 3.5\% of them and writes small policy networks for three MinAtar games from scratch. In three of its experiments the LLM changes a network between a budgeted number of forward passes. It writes a Go network of 1,480,274 weights, plays practice games against fixed opponents, and keeps a change only if it wins more. Its final score comes from 2,000 games against each of two small networks that were trained with proximal policy optimization and that are not the practice opponents. It edits 131,072 weights of a small language model on practice problems, which with 50 forward passes raises a mathematics benchmark score from 7.9\% to 48.1\%. The same procedure did not improve a second, smaller language model.

Several reports followed within days of that post. In WeightLoop~\cite{WeightLoop} an LLM writes a program only once, and that program keeps editing the parameters of a predictor as new measurements arrive. In the work of Gross~\cite{Gross2026}, an LLM writes every weight of a small policy network for an environment with a single decision maker, which is modeled as a Markov decision process. The weights can then be repaired on a dataset of optimal actions that the model checker Storm computes from the model. The LLM revises them from the model-checking results until a requirement holds or a round limit is reached. For each policy that it checks, Storm computes the probability that the requirement holds in the Markov chain that the policy induces on the model, rather than estimating it from simulations. Each revised network becomes the next one to check without an acceptance test, and the pipeline returns the checked network with the best probability, which need not satisfy the requirement when the round limit is reached. Zheng and Doubao~\cite{Zheng2026} propose, without experiments, a loop that accepts a weight edit only if an image-quality score rises by a threshold. Garrido-Merch\'{a}n~\cite{Garrido2026} has an LLM set every weight of feed-forward networks for a set of two-player games from a specification. Zhang et al.~\cite{Zhang2026} and Hu et al.~\cite{Hu2026} let an LLM edit the weights of robot policies and of language models within an evaluation budget, and the LLM decides whether to keep its edit or the original weights. Nusrat and Nusrat~\cite{Nusrat2026} and Ahmed~\cite{Ahmed2026} study the auditing and the reading of weights. Earlier work, NeuroGen~\cite{NeuroGen} and LLM-HYPER~\cite{LLMHYPER}, generates parameters in one step with a tuned or a prompted LLM. None of these works examines several self-interested players that adapt their policies in turn. Improving a single network over several rounds from its own evaluation results is prior work, in Takehi et al., Gross, Zhang et al.\ and Hu et al.

\subsection{LLMs as Iterative Optimizers}

OPRO~\cite{OPRO} places earlier solutions and their scores into the prompt and asks for a better one. Eureka~\cite{Eureka} and AlphaEvolve~\cite{AlphaEvolve} evolve reward functions and programs, respectively, against an evaluator. In EvoPrompt~\cite{EvoPrompt} and STOP~\cite{STOP} the LLM does not see the scores. They select which outputs are kept and, in EvoPrompt, which prompts the LLM receives as parents. Our feedback loop follows the pattern of OPRO. It differs in what the LLM returns, which is a program that writes the complete weight set of a control policy.

\subsection{LLM-Written Strategies in Games}

Code-Space Response Oracles (CSRO)~\cite{CSRO} replaces the RL component of Policy-Space Response Oracles (PSRO)~\cite{PSRO} that computes a best response with an LLM that writes policies as code. It optionally refines a policy that loses against the current mixture, keeps a refinement only if the utility of the policy rises, and adds each result to a population. Bachrach et al.~\cite{Bachrach2025} run a similar loop over programs for Checkers, in which the LLM is asked for a bot that beats a mixture of the earlier bots. Gallego~\cite{Gallego2026} lets an LLM use self-play feedback to rewrite one program that all agents of a sequential social dilemma run.

Two works are closest to the adoption process of this paper. In Opponent-Aware Selective Evolution (OASE)~\cite{OASE2026}, an editor LLM revises the text skill library of each of four LLM agents in first-price auctions and in Cournot competition. Each agent evaluates several revisions against frozen snapshots of its opponents and keeps those whose estimated payoff gain meets a threshold. It adopts the highest-ranked of these under a score that combines the payoff gain with textual novelty, and the adopted revisions of all agents take effect at the same time. The adoption test of this paper is of the same kind. Here the adopted object is a complete set of network weights, which runs without an LLM, and one UAV changes per attempt against the current weights of the others.

The second work is Programmatic Iterated Best Response (PIBR) by Lin et al.~\cite{LinPIBR2025}. Two agents take turns. The LLM uses unit-test errors and episode trajectories to refine the policy code of the updating agent against the fixed code of the other agent. The refined code replaces the previous code without a test of whether the payoff of the agent rose, and the algorithm returns the profile with the highest social welfare. Its formulation allows a separate reward for each agent, and all of its experiments are cooperative. In its matrix games both agents receive the same payoff, and its foraging task is fully cooperative. PIBR shares with this paper the turn-by-turn update against fixed policies and the return of game feedback to the LLM. It differs in three respects. Its LLM writes the policy code itself, while here the LLM writes a program that sets every weight of a neural policy. Its experiments are cooperative, while here the interests of the UAVs conflict. It has no acceptance test, while here a UAV adopts a candidate only if its own score rises by a margin.

\subsection{Position of This Paper}

Table~\ref{tab:related} summarizes the comparison. This paper differs from the inspected sources in the combination of three properties. The LLM writes a program that sets every weight of a neural control policy, and it may read but not execute the simulator. The LLM receives the evaluation results and earlier programs, and in the game loops also the current programs of the other UAVs. A new weight set is adopted unilaterally by one UAV on the basis of that UAV's own score. The third property is the subject of this paper. Its rule follows game theory. A sequence of profiles in which one player at a time changes its strategy and raises its own payoff is an improvement path in the sense of Monderer and Shapley~\cite{MondererShapley1996}, and OASE applies a payoff threshold of the same kind to the text strategies of LLM agents. What is new here is the source of the candidates and the adopted object. The candidates are programs that an LLM writes, and the adopted object is a complete set of network weights. In unilateral adoption an attempt costs one LLM answer and 10 evaluation episodes, and the adopted weights run without LLM calls. We did not find a source in which several self-interested players of a game adopt LLM-written network weights in turn, each on the basis of its own payoff. Among the works on LLM-written weights, the nearest case is the Go experiment of Takehi et al., in which the LLM keeps a change to a single network only if the network wins more against fixed opponents. This statement is limited to the search that Appendix~\ref{appx:search} records.

\begin{table*}[t]
\centering
\caption{Closest related works. ``Feedback'' states whether evaluation results return to the LLM between its outputs. The table lists works in which an LLM writes policies, weights, programs or prompts. Works in which an LLM only sets the parameters of a game~\cite{Tang2026, Nguyen2026} are not listed.}
\label{tab:related}
\footnotesize
\begin{tabularx}{\textwidth}{>{\raggedright\arraybackslash}p{3.4cm}>{\raggedright\arraybackslash}p{4.0cm}>{\raggedright\arraybackslash}p{4.6cm}X}
\toprule
Work & LLM output & Feedback & How a new policy or edit is taken up \\
\midrule
Takehi et al.~\cite{Takehi2026} & weight values, or code that constructs them & yes, in the Go and language-model editing experiments & single network, in Go a change is kept if it wins more \\
WeightLoop~\cite{WeightLoop} & a parameter-editing program that the LLM writes once & the program iterates, the LLM does not & single predictor \\
Gross~\cite{Gross2026} & weight values & yes, states that miss an optimal action (optional step) and model-checking results & single policy, stops when a requirement holds or at a round limit and returns the best checked network \\
Zheng and Doubao~\cite{Zheng2026} & weight edits (proposal) & proposed & single model, threshold on a quality score \\
Garrido-Merch\'{a}n~\cite{Garrido2026} & weights & during construction & two-player games, no adoption \\
Zhang et al.~\cite{Zhang2026}, Hu et al.~\cite{Hu2026} & weight edits & yes & single model, the editing LLM keeps or reverts its edit \\
Nusrat and Nusrat~\cite{Nusrat2026}, Ahmed~\cite{Ahmed2026} & weight edits, or a reading of weights & no & single network \\
NeuroGen~\cite{NeuroGen}, LLM-HYPER~\cite{LLMHYPER} & parameters & no (NeuroGen uses the task loss only to train its generator) & single model \\
OPRO~\cite{OPRO}, Eureka~\cite{Eureka}, AlphaEvolve~\cite{AlphaEvolve} & prompts or code & yes & single task \\
EvoPrompt~\cite{EvoPrompt}, STOP~\cite{STOP} & prompts or code & no, scores only select outputs (and, in EvoPrompt, parents) & single task \\
CSRO~\cite{CSRO}, Bachrach et al.~\cite{Bachrach2025} & strategy code & utility of a losing policy against the current mixture (CSRO), code of a mixture bot that samples an earlier bot before each game from an approximate Nash mixture of the estimated payoff matrix (Bachrach et al.) & added to a population \\
Gallego~\cite{Gallego2026} & strategy code & yes, from self-play & one program shared by all agents \\
OASE~\cite{OASE2026} & revisions of a text skill library & yes, a memory of the agent's own outcomes & each agent adopts a revision whose estimated payoff gain meets a threshold, all agents at once \\
Lin et al.~\cite{LinPIBR2025} (PIBR) & policy code & yes, unit-test errors and episode trajectories & agents in turn, cooperative tasks, no acceptance test \\
This paper & a program that sets every weight of a policy & yes, scores, metrics, earlier programs and, in the game loops, the current programs of the other UAVs & one UAV at a time adopts on the basis of its own score, and the adopted weights run without an LLM. Imposed fleet-wide replacement in separate runs \\
\bottomrule
\end{tabularx}
\end{table*}

\section{System Model}
\label{sec:system}

This section describes the simulated system with 30 UAVs and static users, which we call the 30-UAV configuration. It is used by the runs in which the LLM writes one weight set for all 30 UAVs (the single-fleet setting of Section~\ref{sec:problem}), by the RL baselines and by the 30-UAV game loop. Section~\ref{sec:gameloops} describes the 5-UAV configuration. Table~\ref{tab:settings} lists the main settings with their sources and marks the ones that we chose. None of the choices that Table~\ref{tab:settings} marks as ours was calibrated or compared with alternatives. Appendix~\ref{appx:sim} gives the remaining details of the simulator.

\subsection{Area, UAVs and Worlds}
\label{sec:area}

As in Yang's thesis~\cite{YangThesis}, the area is a 4~km by 4~km disaster zone that is divided into 20 by 20 cells of 200~m. Thirty UAVs act as aerial base stations. UAVs 0 to 17 are of Yang's first type, and UAVs 18 to 29 are of his second type. The two types differ in battery capacity, in power draw and in the altitudes that they may use. A UAV of the first type chooses its altitude between 50 and 300~m, and a UAV of the second type flies at 50~m. The lower bound is the altitude of Yang's low-altitude mode. We chose 300~m as the upper bound. It is the maximum target height that 3GPP TR~36.777~\cite{TR36777} considers for aerial user equipment, which we use as a numerical reference and not as a requirement for base stations. One step lasts 10~s, and one episode lasts $T = 1000$ steps.

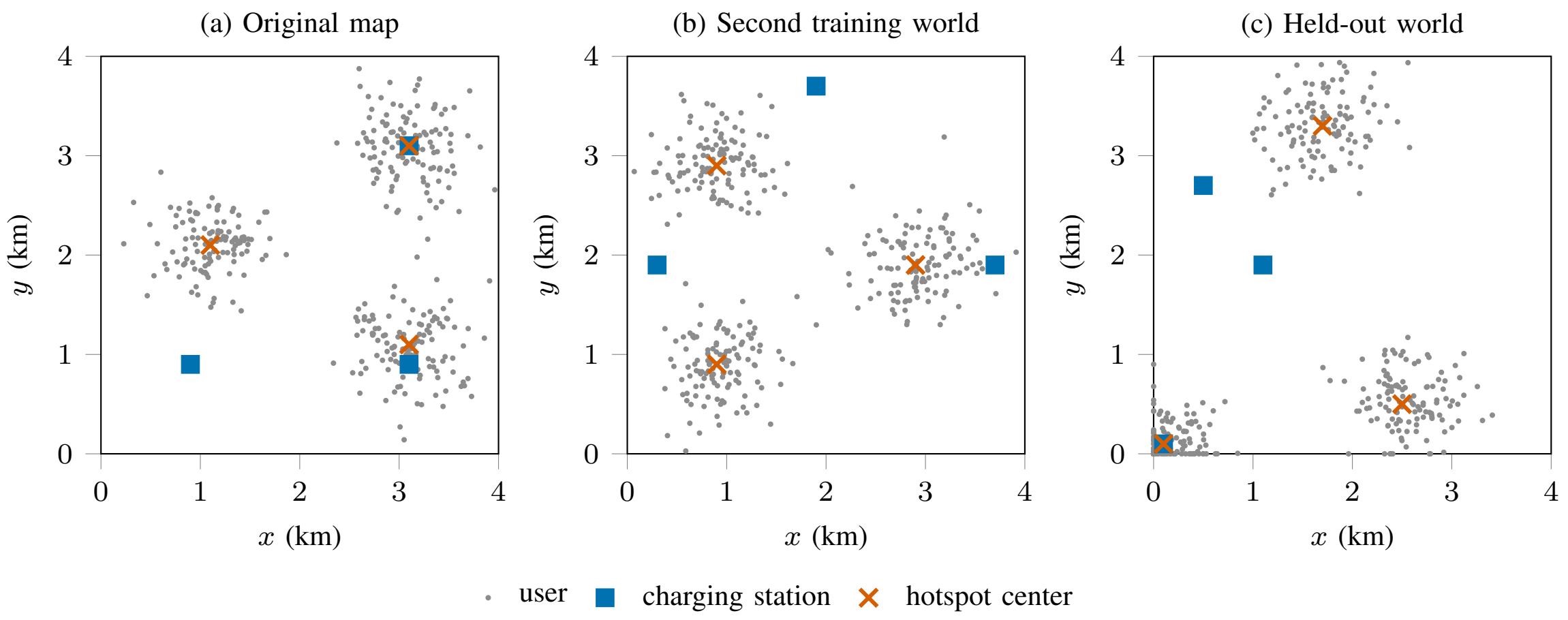
\begin{figure*}[t]
\centering
\begin{tikzpicture}
\begin{groupplot}[
  group style={group size=3 by 1, horizontal sep=1.25cm},
  width=0.30\textwidth, height=0.30\textwidth,
  xmin=0, xmax=4, ymin=0, ymax=4, xtick={0,1,2,3,4}, ytick={0,1,2,3,4},
  xlabel={$x$ (km)}, ylabel={$y$ (km)},
  tick label style={font=\footnotesize}, label style={font=\footnotesize}, title style={font=\footnotesize, yshift=-1ex},
  axis equal image, tick align=outside, tick pos=left,
  legend style={font=\footnotesize, draw=none, legend columns=3, column sep=6pt},
]
\nextgroupplot[title={(a) Original map}, legend to name=scenariolegend]
\addplot[only marks, mark=*, mark size=0.55pt, black!45] table[x=x, y=y]{figs/data/world-original.dat};
\addlegendentry{user}
\addplot[only marks, mark=square*, mark size=2.4pt, cBlue] coordinates {(3.1,3.1) (0.9,0.9) (3.1,0.9)};
\addlegendentry{charging station}
\addplot[only marks, mark=x, mark size=3.4pt, cRed, line width=1.1pt] coordinates {(1.1,2.1) (3.1,1.1) (3.1,3.1)};
\addlegendentry{hotspot center}
\nextgroupplot[title={(b) Second training world}]
\addplot[only marks, mark=*, mark size=0.55pt, black!45] table[x=x, y=y]{figs/data/world-second.dat};
\addplot[only marks, mark=square*, mark size=2.4pt, cBlue] coordinates {(1.9,3.7) (3.7,1.9) (0.3,1.9)};
\addplot[only marks, mark=x, mark size=3.4pt, cRed, line width=1.1pt] coordinates {(2.9,1.9) (0.9,2.9) (0.9,0.9)};
\nextgroupplot[title={(c) Held-out world}]
\addplot[only marks, mark=*, mark size=0.55pt, black!45] table[x=x, y=y]{figs/data/world-heldout.dat};
\addplot[only marks, mark=square*, mark size=2.4pt, cBlue] coordinates {(0.1,0.1) (0.5,2.7) (1.1,1.9)};
\addplot[only marks, mark=x, mark size=3.4pt, cRed, line width=1.1pt] coordinates {(0.1,0.1) (2.5,0.5) (1.7,3.3)};
\end{groupplot}
\end{tikzpicture}\\[0.5ex]
\pgfplotslegendfromname{scenariolegend}
\caption{The three worlds of the 30-UAV configuration. Each world fixes the positions of 360 users in three clusters, of three charging stations and of three hotspot centers, and the channel of every UAV (not shown). In (a) and (c), one station lies on a hotspot center.}
\label{fig:worlds}
\end{figure*}

A \emph{world} is a scenario file that fixes the cells of the three charging stations, the three hotspot centers, the positions of the users (360 in the 30-UAV configuration) and the channel of every UAV. The station and hotspot cells are inputs of a world, and Fig.~\ref{fig:worlds} shows them. A world seed then determines the user positions and the channel assignment. In the experiments that this paper analyzes, the users do not move. The users form three Gaussian clusters of 120 users around the hotspot centers, with a standard deviation of 300~m, and positions are clipped to the area. The user count follows Chou and Yen~\cite{ChouYen}. We chose the spread after inspecting the clustering experiments in their Fig.~7. Fig.~\ref{fig:worlds} shows the three worlds of this paper, which were generated with world seeds 17, 18 and 19. The first world uses Yang's station and hotspot positions, and we call it the original map. Yang's demand map is empty in the 3 by 3 cells around each station, whereas our user model has no demand-free zone around the stations. The second world uses station and hotspot cells of our own choice. The original map and the second world are the two \emph{training worlds}. The third world is a \emph{held-out world} whose stations and hotspots lie at positions that neither training world uses. Its station and hotspot cells were drawn by a script from a random order of the cells. The center of each station cell keeps at least 800~m from the center of every station cell of the training worlds, and each hotspot keeps the same distance from every hotspot of the training worlds. The value of 800~m is the default of the script and is stated in our record of the generation. The seed of the random order was not recorded, and a draw with seed 18 reproduces the layout exactly.

At the start of an episode, UAV $i$ is in flight at an altitude of 50~m above station $i \bmod 3$. Before the first action, no UAV transmits and no user is connected. The initial battery of UAV $i$ is drawn uniformly from one of three bands, 10\% to 20\%, 45\% to 55\% or 80\% to 90\% of its capacity, according to its index. The simulator does not model collisions, so several UAVs may occupy the same position. The bands follow Yang, and they make every episode start with some UAVs close to needing a charge. An evaluation consists of 5 episodes per world with fixed episode seeds. The episode seed changes only the battery draws. The 5 episodes of a world are therefore 5 initializations of the same world and not 5 independent worlds.

\subsection{Energy and Charging}
\label{sec:energy}

Energy is counted in joules per step and computed from Yang's power constants (Table~\ref{tab:energy} in Appendix~\ref{appx:sim}). A UAV that moves uses flight energy in proportion to its horizontal distance and climb or descent energy in proportion to its change of altitude, and it uses at least the hover energy. A UAV that serves uses the larger of this movement energy and the service energy. A UAV whose battery reaches zero crashes. It is removed for the rest of the episode and earns nothing afterward.

Each of the three charging stations has 5 charging slots and 5 waiting slots, and the stations apply Yang's queue and departure rules. A UAV that arrives takes a charging slot if one is free and a waiting slot otherwise. A station whose slots are all taken rejects the UAV, which then hovers and targets another station. UAVs that arrive in the same step compete for the slots before departures of that step release capacity. They are sorted by the battery fraction after the flight and descent energy, in units of $10^{-6}$ and rounded down, and a lower value comes first. Ties favor a UAV that returns to its last station, and the order of arrivals that are equal in both respects is not specified. When a charging slot becomes free, a waiting UAV whose battery is nearly empty takes it first (Appendix~\ref{appx:sim}). Otherwise the waiting UAV with the largest priority $0.95\,(1 - s) + 0.05\,q$ takes it. Here $s$ is the state of charge of the UAV, that is, its battery fraction, and its service efficiency $q$ is the service component of its reward (Section~\ref{sec:reward}) per step from its last departure, or from the start of the episode, until its arrival at the station. The value of $q$ is clipped to $[0, 1]$. The service component of this paper thus also enters the charging priority and the departure threshold through $q$. A charging UAV leaves when its state of charge reaches $0.80 + 0.05\,q$ or when the station forces it out. While a UAV is at a station, its policy has no effect.

\subsection{Radio and Association}
\label{sec:radio}

The radio model follows Chou and Yen~\cite{ChouYen} with the changes that Table~\ref{tab:settings} marks: frequency channels, thermal noise, a handover margin, transmission only in serve mode, a different path-loss model, a lower transmit power, altitudes chosen by the UAVs, and a cap on the SINR in the score.

There are 10 frequency channels of 1~MHz, and three UAVs share each channel. The channel $\kappa_j$ of UAV $j$ is drawn once per world and does not change. Consider a UAV at altitude $z$ whose horizontal distance to a user is $d_h$. The user antenna is at height $h_u = 1.5$~m. The slant distance is $d = \max\!\big(1\ \mathrm{m}, \sqrt{d_h^2 + (z - h_u)^2}\big)$, and the elevation angle in degrees is $\phi = (180/\pi)\operatorname{atan2}(z - h_u, d_h)$, where $\operatorname{atan2}$ is the two-argument arctangent. The received power in dBm is
\begin{equation}
\begin{aligned}
P^{\mathrm{dBm}}_{\mathrm{rx}} &= P^{\mathrm{dBm}}_{\mathrm{tx}} - \mathrm{PL}(d, \phi), \\
\mathrm{PL}(d, \phi) &= 20\log_{10}\!\Big(\frac{4\pi f_{\mathrm{c}} d}{c}\Big) + \eta_{\mathrm{LoS}}\,p(\phi) \\
&\quad + \eta_{\mathrm{NLoS}}\,\big(1 - p(\phi)\big),
\end{aligned}
\label{eq:pl}
\end{equation}
where $P^{\mathrm{dBm}}_{\mathrm{tx}} = 30$~dBm is the transmit power, $f_{\mathrm{c}} = 2$~GHz is the carrier frequency, $c$ is the speed of light, and
\begin{equation}
p(\phi) = \frac{1}{1 + a\,\exp\!\big(-b\,(\phi - a)\big)}
\label{eq:plos}
\end{equation}
is the line-of-sight (LoS) probability of Al-Hourani et al.~\cite{AlHourani2014}. The simulator uses the urban values $a = 9.61$ and $b = 0.16$ and the excess losses $\eta_{\mathrm{LoS}} = 1$~dB and $\eta_{\mathrm{NLoS}} = 20$~dB under LoS and non-line-of-sight (NLoS) conditions. The path loss of Eq.~\eqref{eq:pl} is thus an average over the LoS probability, and the simulator does not draw a LoS state for a link. These four values are listed in Table~I of Bor-Yaliniz et al.~\cite{BorYaliniz2016}, who state that they calculated them from~\cite{AlHourani2014} and Recommendation ITU-R P.1410. El Hammouti et al.~\cite{ElHammouti2020} use the same values.

Powers are converted to milliwatts by $P^{\mathrm{mW}} = 10^{P^{\mathrm{dBm}}/10}$ before they are combined. Only UAVs in serve mode (Section~\ref{sec:modes}) transmit. For a user $u$ and a transmitting UAV $j$,
\begin{equation}
\mathrm{SINR}_{u,j} = \frac{P^{\mathrm{mW}}_{\mathrm{rx}}(u,j)}{\displaystyle\sum_{\substack{l \ne j,\ \kappa_l = \kappa_j \\ l\ \mathrm{transmitting}}} P^{\mathrm{mW}}_{\mathrm{rx}}(u,l) + N_0}.
\label{eq:sinr}
\end{equation}
The noise power $N_0 = 10^{-10.5}$~mW, which is $-105$~dBm, follows from the thermal noise density of $-174$~dBm/Hz at 290~K~\cite{ITUR1753}, the bandwidth of 1~MHz and the 9~dB noise figure that TR~36.777 gives for user equipment.

UAV $j$ is \emph{eligible} for user $u$ if it transmits, $P^{\mathrm{dBm}}_{\mathrm{rx}}(u,j) \ge -100$~dBm and $\mathrm{SINR}_{u,j} \ge 0.1$ (that is, $-10$~dB). These thresholds follow Chou and Yen. A user that is not connected compares the eligible UAVs on all channels and connects to the one with the highest SINR. Equal highest SINRs favor the UAV with the lowest index. A connected user keeps its UAV while that UAV stays eligible, unless another eligible UAV offers an SINR that is at least 2~dB higher. When its UAV is no longer eligible, the user selects again. A change from one UAV to another counts as a handover. We use the margin to avoid repeated switching between two UAVs that offer similar SINRs. Its value of 2~dB is the A3 offset in the mobility assumptions of TR~36.777~\cite{TR36777}, which we apply to SINR differences without the 160~ms time to trigger. The simulator does not model admission control, so the network never rejects a connection, and it does not model the link from a UAV to the core network. With these simplifications, users switch among the eligible UAVs on all channels without admission limits or a backhaul constraint.

\subsection{Modes, Movement and the Energy Guard}
\label{sec:modes}

At every step, each UAV chooses one of three modes. In \emph{move} mode the UAV flies, does not transmit and causes no interference. In \emph{serve} mode the UAV transmits, serves its connected users and may also move. In \emph{charge} mode the UAV flies straight to the nearest station that has not rejected it and charges there. We chose the rule that only UAVs in serve mode transmit so that a UAV can reposition without interfering. The model of Chou and Yen has no such mode, and every UAV contributes to the interference there. Our rule therefore changes eligibility, interference, energy use and reward together.

In move and serve modes, the movement is the 3-dimensional output of the policy. The movement head is linear, without an output activation. Its output is scaled by (200~m, 200~m, 50~m) and limited to a length of 200~m per step. The position is kept inside the area and inside the altitude range of the UAV's type. The charge mode is \emph{selectable} only when the nearest station is within 100~m and has a free charging or waiting slot, or when the UAV is already in charge mode. Otherwise a charge request counts as a move.

The \emph{energy guard} is a rule that we added for some RL training runs and for the evaluations in which the guard is on. Its reserve $R$ is an estimate that is computed from distances. It is the flight energy for the horizontal distance to the nearest station, plus the flight energy for twice the distance from that station to its nearest neighboring station, plus the descent energy from the current altitude. It does not include the hover energy of the steps on the way or the energy used while waiting. The factor of two is our choice and was not compared with other values in this simulator. The guard does not guarantee survival, and Table~\ref{tab:fleet} shows crashes with the guard on. When the battery of a UAV falls below $R$, the guard forces charge mode, also outside the region in which charge mode is selectable, and keeps it active until the UAV departs from a station. The guard is off in every evaluation that selects a candidate and in every game loop.

\subsection{Observation and Policy Network}
\label{sec:obs}

Each UAV observes 880 numbers at every step. They consist of a 20 by 20 grid of 90~m cells centered on the UAV with the number of users and the number of its own connected users in each cell (800 numbers); 6 numbers for each of up to 8 other UAVs inside this 1.8~km window (48); the number of its connected users (1); its own position (3); Yang's 19 per-UAV scalars, which include the battery fraction and the last station visited (19); and the offsets and distances of the three stations (9). Appendix~\ref{appx:sim} gives the order and the scaling. The simulator provides these numbers to every UAV in every mode, and the cost of obtaining them is not modeled.

The policy network has two hidden layers of 256 units with tanh activations and two heads: a 3-dimensional movement vector and 3 logits for the mode. The charge logit is masked when charging is not selectable. At evaluation time the policy acts deterministically. The movement is the network output, and the mode is the one with the largest masked logit, with ties resolved in the order move, serve, charge. In this paper the weights of this network are written by programs that an LLM returns, without gradient updates. The RL baselines train the same architecture with MAPPO.

\subsection{Reward and Reported Score}
\label{sec:reward}

The \emph{service component} of UAV $i$ at step $t$ is
\begin{equation}
v_i(t) = \sum_{u \in U_i(t)} \log_2\!\big(1 + \min(\mathrm{SINR}_{u,i}(t), 10^4)\big),
\label{eq:service}
\end{equation}
where $U_i(t)$ is the set of users connected to UAV $i$ at step $t$. It follows Eqs.~(3)--(5) of~\cite{ChouYen} and sums the spectral efficiencies of the links of the UAV, with the SINR capped at $10^4$ (40~dB). It does not divide resources among the users of one UAV, so it is not a delivered throughput. The per-step reward of UAV $i$ is
\begin{equation}
r_i(t) =
\begin{cases}
v_i(t) + c_i(t), & \text{if UAV $i$ is alive after step $t$,}\\
-6, & \text{if UAV $i$ crashes in step $t$,}\\
0, & \text{if UAV $i$ crashed before step $t$.}
\end{cases}
\label{eq:reward}
\end{equation}
The charging term $c_i(t)$ and the crash reward of $-6$ are taken from Yang. The charging term equals $6\,\big(1 - 2/(1 + e^{-6.5\,(s - 0.5)})\big)$ on the first step of each station visit in which the station accepts the UAV, in a charging or a waiting slot, and zero on every other step. Here $s$ is the battery fraction after the energy changes of that step. The term is not paid on the last step of an episode. The term is positive below a battery fraction of 0.5 and negative above it, and entry into a waiting slot also receives it. The crash reward replaces the service component and the charging term of the step in which the UAV crashes.

The \emph{score} of a UAV in an episode is the sum of its per-step rewards divided by $T$. The \emph{fleet-mean score} is the mean of the scores of all UAVs, and this mean includes the UAVs that crashed. Unless stated otherwise, reported values are means over the 5 evaluation episodes of a world. The score has no physical unit, and its service part is a sum of spectral efficiencies in bit/s/Hz.

\begin{table*}[t]
\centering
\caption{Settings of the 30-UAV configuration and their sources. ``This paper'' marks our own implementation choices. The stated reasons explain their intended roles. Their numerical values were not calibrated or compared with alternatives.}
\label{tab:settings}
\footnotesize
\begin{tabularx}{\textwidth}{>{\raggedright\arraybackslash}p{2.4cm}>{\raggedright\arraybackslash}p{6.9cm}X}
\toprule
Setting & Value & Source \\
\midrule
Area, step, episode & 4~km by 4~km, cells of 200~m, 10~s per step, 1000 steps, at most 200~m of movement per step & Yang~\cite{YangThesis} \\
UAVs & 30, two types (UAVs 0--17 and 18--29) & Yang \\
Altitude & 50 to 300~m (first type), 50~m (second type) & 50~m: Yang. 300~m: limit for aerial user equipment in TR~36.777~\cite{TR36777}. Range: this paper \\
Energy & constants of Table~\ref{tab:energy}, crash reward $-6$ & Yang, Tables~6.2 and~6.3 and his code \\
Stations & 3 stations with 5 charging and 5 waiting slots, Yang's queue and departure rules & Yang. Selectability radius of 100~m, handling of rejections and straight flight to the station: this paper \\
Users & 360 static users in 3 Gaussian clusters of 120, standard deviation 300~m & Count: Chou and Yen~\cite{ChouYen}. Spread: this paper \\
Channels & 10 channels of 1~MHz, 3 UAVs per channel, fixed per world & This paper. It limits the interferers of each UAV to two \\
Noise & $-105$~dBm & Noise figure of TR~36.777, 1~MHz, thermal density~\cite{ITUR1753} \\
Path loss & Eqs.~\eqref{eq:pl} and~\eqref{eq:plos} at 2~GHz with urban parameters & Model: Al-Hourani et al.~\cite{AlHourani2014}. Values as in Table~I of~\cite{BorYaliniz2016} \\
Transmit power, user antenna height & 30~dBm, 1.5~m & This paper. 1.5~m is the lowest user height in the channel models of TR~36.777 \\
Association & eligibility at $-100$~dBm and SINR 0.1, best SINR on first connection, 2~dB handover margin & Thresholds: Chou and Yen. Margin: A3 offset of TR~36.777, applied to SINR differences \\
Transmission & only UAVs in serve mode transmit & This paper \\
Reward & Eqs.~\eqref{eq:service} and~\eqref{eq:reward} & Service: Eqs.~(3)--(5) of~\cite{ChouYen}. Charging and crash terms: Yang. Cap: this paper \\
Observation & 880 numbers, window of 1.8~km & Layout: this paper. The 1.8~km window keeps the range of Yang's 9 by 9 view of 200~m cells. The 19 per-UAV scalars: Yang's code \\
Network & 880--256--256 (tanh), movement head and 3 mode logits & This paper \\
\bottomrule
\end{tabularx}
\end{table*}

\section{Problem Formulation}
\label{sec:problem}

\subsection{Players, Policies and Payoffs}

Table~\ref{tab:notation} lists the main notation. Let $N$ be the set of UAVs. UAV $i \in N$ runs the policy network of Section~\ref{sec:obs} with its own weight set $\theta_i$, and a \emph{profile} is the tuple $(\theta_i)_{i \in N}$. The score of UAV $i$ in one episode is
\begin{equation}
J_i(\theta_i, \theta_{-i}) = \frac{1}{T}\sum_{t=1}^{T} r_i(t),
\label{eq:own}
\end{equation}
where $\theta_{-i}$ denotes the weight sets of the other UAVs and $r_i(t)$ is the reward of Eq.~\eqref{eq:reward}. The score of one UAV depends on the others through the environment. Same-channel UAVs that transmit near the users of UAV $i$ lower the SINR of those users. Every user connects to the eligible UAV with the best SINR under the handover rule of Section~\ref{sec:radio}, so UAVs take users from each other. The UAVs also share the slots of the charging stations. The fleet-mean score is
\begin{equation}
J = \frac{1}{|N|}\sum_{i \in N} J_i .
\label{eq:fleet}
\end{equation}
The sum $|N|\,J$ of the scores corresponds to the social welfare in the game of Chou and Yen~\cite{ChouYen}, so the fleet-mean score is the social welfare per UAV.

We study two settings. In the \emph{single-fleet} setting all UAVs share one weight set $\theta$, and we rank LLM-written candidates by the fleet mean of the evaluation payoffs that are defined below. Yang's thesis is of this kind: all UAVs execute one shared policy, and the thesis states a common system-level objective. In his code the shared policy is trained on the reward of each UAV, which includes a global coordination term. Our MAPPO baselines also share the actor parameters and train on each UAV's own reward, which has no such term (Appendix~\ref{appx:rl}). In the \emph{game} setting each UAV is a player that cares only about its own score $J_i$. The fleet-mean score is still reported, but no player maximizes it. Apart from the zero-crash condition of the single-fleet runs (Section~\ref{sec:oneshot}), a crash is not forbidden in either setting. A crash gives the UAV a reward of $-6$ in that step and removes all of its later rewards, and in the game setting it enters an adoption decision only through the score of the focal UAV. Selection and adoption decisions in both settings use the \emph{evaluation payoff} $\bar J_i$ and its fleet mean $\bar J$, which are means over 5 evaluation episodes on each world that a run uses. Table~\ref{tab:loops} and Appendix~\ref{appx:repro} state which worlds each run uses.

\begin{table}[t]
\centering
\caption{Main notation.}
\label{tab:notation}
\footnotesize
\begin{tabularx}{\columnwidth}{>{\raggedright\arraybackslash}p{1.6cm}X}
\toprule
Symbol & Meaning \\
\midrule
$N$, $T$ & set of UAVs, number of steps in an episode \\
$\kappa_j$ & frequency channel of UAV $j$ \\
$d$, $\phi$ & slant distance and elevation angle between a UAV and a user \\
$N_0$ & noise power \\
$s$, $q$ & state of charge and service efficiency of a UAV \\
$R$ & reserve of the energy guard \\
$U_i(t)$ & set of users connected to UAV $i$ at step $t$ \\
$r_i(t)$, $v_i(t)$, $c_i(t)$ & reward, service component and charging term of UAV $i$ at step $t$ \\
$\theta_i$, $\theta_{-i}$, $\theta$ & weight set of UAV $i$, weight sets of the other UAVs, shared weight set \\
$J_i$, $J$ & score of UAV $i$ in an episode, fleet-mean score \\
$\bar J_i$, $\bar J$ & evaluation payoff of UAV $i$, its fleet mean \\
$i_k$, $\theta'$ & focal UAV of attempt $k$, candidate weight set \\
$\Delta$, $\epsilon$ & change in the payoff of the focal UAV, adoption margin \\
$\tau$ & battery threshold of the penalty in two RL baselines \\
\bottomrule
\end{tabularx}
\end{table}

\subsection{Unilateral Adoption}
\label{sec:adoption}

A profile changes through the following process, which runs in \emph{attempts}. In attempt $k$ one UAV $i_k$ is the \emph{focal UAV}, and one candidate weight set $\theta'$ is written for it and tested. The candidate is evaluated while every other UAV keeps its current weight set, and the change in the payoff of the focal UAV is
\begin{equation}
\Delta = \bar J_{i_k}(\theta', \theta_{-i_k}) - \bar J_{i_k}(\theta_{i_k}, \theta_{-i_k}).
\end{equation}
The focal UAV adopts the candidate if
\begin{equation}
\Delta > 0 \quad\text{and}\quad \Delta \ge \epsilon\,\big\lvert \bar J_{i_k}(\theta_{i_k}, \theta_{-i_k}) \big\rvert ,
\label{eq:adopt}
\end{equation}
and in that case only $\theta_{i_k}$ changes. The adoption margin is $\epsilon = 0.01$ in every run that this paper analyzes. We chose this value. It was not calibrated against the variation of the score between evaluation episodes, and the adoption and stopping results that this paper analyzes depend on it. Focal UAVs are visited in a fixed cyclic order that starts with UAV~0. After an adoption, the scores of all UAVs under the new profile become the reference for later attempts. The process stops when every UAV has had a candidate rejected since the last adoption, when a cap on the number of attempts is reached, or after five technical failures in a row (Section~\ref{sec:gameloops}). A technical failure does not count as a rejection. Without technical failures the first condition is one full cycle without an adoption. At such a stop every UAV has had one candidate rejected, and we did not test whether the final profile is an $\epsilon$-equilibrium. The 30-UAV replacement run of Table~\ref{tab:loops} was set to a different rule, 10 consecutive attempts without an adoption. In that run 79 of 100 candidates were adopted, and from attempt 37 on every cycle of 8 attempts contained adoptions, so the run met neither rule (Section~\ref{sec:games}).

The other UAVs keep their weight sets during an attempt, but they still act on their observations. Their policies are fixed within an evaluation and may change in later attempts in which they are focal.

In the terms of Monderer and Shapley~\cite{MondererShapley1996}, the adopted profiles form an improvement path with respect to the evaluation payoffs, because at each adoption a single player changes its strategy and its own evaluation payoff strictly rises. Their $\epsilon$-improvement path requires a gain above an absolute $\epsilon$, whereas Eq.~\eqref{eq:adopt} uses a margin relative to the current payoff, and each attempt tests a single candidate. What is specific to this paper is the source of the candidate, which is a program that an LLM writes and that sets every weight of a neural policy. The candidate is not computed as a best response, exact or approximate. When the process stops by the first condition, each UAV has had one candidate rejected against the final profile. A stop at the cap or after technical failures does not give even this. In all cases nothing is known about untested weight sets, so the final profile is not shown to be a Nash equilibrium or an $\epsilon$-equilibrium. We make no claim that this game has a potential function, and convergence of the policies is not established.

\subsection{Imposed Fleet-Wide Replacement}
\label{sec:replacement}

We also study \emph{imposed fleet-wide replacement}, a procedure in which the fleet shares one weight set. A candidate is written for the focal UAV and is evaluated with the focal UAV alone running it. If Eq.~\eqref{eq:adopt} holds for the focal UAV, the experiment imposes the candidate on every UAV. The adoption decisions of the other players are not modeled, so this variant tests a procedure and not a decision rule of the players. It uses the gain that one UAV achieves against a fixed fleet to decide a change of the whole fleet. The two procedures differ in two ways at once: whether one UAV or the whole fleet changes, and whether the weight sets are individual or shared. Our experiments do not separate these two differences. In the main text, unilateral adoption is analyzed in a run with 5 UAVs and imposed fleet-wide replacement in a run with 30 UAVs. Appendix~\ref{appx:repro} reports a replacement run with 5 UAVs for a comparison at the same fleet size.

\subsection{Scope of the Game Setting}
\label{sec:scope}

The game of this paper is defined by the simulator and by separate scores of the UAVs. We assume that the UAVs over one disaster area belong to different operators and that each operator cares only about the score of its own UAV. Adoption depends only on the focal UAV's own score. The construction of a candidate is offline and uses the simulator source and the programs that set the current weights of the other UAVs (Section~\ref{sec:gameloops}), so the focal UAV knows the current policies of the others, as the player that changes its strategy does in turn-by-turn improvement in a game. Two things are not established for such operators. The payoff is the simulator reward that we inherited, and its relation to delivered service, to the benefit of an operator and to the cost of a lost aircraft is not established. The paper also does not say who would arrange the shared channels and charging stations.

\section{LLM-Written Weights and the Game Loops}
\label{sec:approach}

The approach has three parts. An LLM writes the complete weight set of the policy network in one answer, which we call \emph{one-shot} weight writing. A feedback loop repeats this task with the evaluation results of earlier rounds. The game loops place the task inside the adoption process of Section~\ref{sec:problem}. The ancillary file task-texts.txt of this preprint reproduces the single-fleet task and the game-specific sections of two game tasks.

\subsection{Writing the Weights}
\label{sec:oneshot}

The LLM receives a written task. It describes the system of Section~\ref{sec:system} in plain words, the score to maximize, the order and scaling of the 880 inputs, the network and the names of its weight tensors, a reference weight set, the condition of the run, the way the answer is scored, and the answer format. The task permits read-only inspection of files. It prohibits running an interpreter and executing the environment, the model or the evaluator. We chose this restriction because a player in a game is assumed to reason from the rationality of the others and of itself rather than to try out every strategy. The session logs contain no command that does either (Appendix~\ref{appx:repro}). The LLM is OpenAI's GPT-6.1 Sol. We call it through the Codex command-line tool with the reasoning effort \texttt{xhigh} and a read-only sandbox. The same model and settings were used in every run.

The LLM answers with a short explanation, one Python code block that defines a function \texttt{patch(sd)}, and its own prediction of the score. The function receives the state dictionary of the network, that is, the dictionary of its named weight tensors, and writes the nine tensors of the policy: the two hidden layers with their biases, the movement head and the mode head with their biases, and a vector of log standard deviations that the deterministic evaluation does not use. A candidate is valid if all of the following hold: the answer contains exactly one Python code block; the program passes a keyword filter that rejects selected system, file and network operations; the program applies without error; the tensors keep their names, shapes and data types and are finite; and the evaluation completes. Every returned program that we inspected first sets all nine tensors to zero and then writes them, but the validity check does not enforce this. The keyword filter is not a complete isolation of the execution. In the rest of the paper the returned function is a program, the weights that it sets are a weight set, which is the policy of a UAV, and a weight set that is tested for adoption is a candidate.

According to the task text and our run records, the reference weight set is the actor network of an earlier MAPPO policy, that is, the network that selects the actions. That policy was trained for a previous version of the problem, which had one shared channel, no noise, no handover margin, no choice between move and serve, and 111 inputs. The task gives it as one text file per tensor. It maps 31 of the 111 inputs to inputs 849 to 879 of the present network, which describe the position of the UAV, its 19 scalar features and the 9 station features. The other 849 inputs of the present network have no counterpart, and the earlier policy has one charge output where the present network has a mode head. The task allows the LLM to reuse, transplant or ignore any part of the reference weight set. The task states that this policy never chose to charge without the energy guard. No returned program that we inspected reads the reference files.

The task is run under two conditions. In the \emph{crash-allowed} condition only the fleet-mean score counts. In the \emph{zero-crash} condition a weight set counts only if no UAV crashes on either training world. A third task, which we call \emph{self-interested}, was run once and in one answer. Its text states that all 30 UAVs run the returned weights and that each UAV belongs to a different operator and counts only its own reward. It asks for weights such that no single UAV could raise its own reward by switching to other weights while the other 29 keep them. The returned weights do not have this property. In the further replacement loop that started from them (Appendix~\ref{appx:repro}), the third attempt found a candidate that raised the payoff of one UAV by 11.8\% while the other 29 UAVs kept the returned weights. The text also says that the answer is not scored directly and becomes the first shared weight set of a replacement loop (Appendix~\ref{appx:repro}).

The sandbox prevents writes but does not confine reads to the working directory. In the single-fleet runs the LLM could read the simulator source, including the energy and charging logic, and the session logs show that it also read files outside the working directory. Appendix~\ref{appx:repro} reports what the logs show.

\subsection{Evaluation-Feedback Loop}
\label{sec:loop}

The loop repeats the task. After each round the returned program is applied to a template state dictionary, and the resulting weights are evaluated for 5 episodes on each of the two training worlds with the energy guard off. The round's fleet-mean score, the per-world metrics, the LLM's explanation and program, and a summary of one episode per world are appended to a history file. The next prompt points the LLM to this history, marks the best round so far and allows it to start from the program of any earlier round. The loop stops after 5 consecutive rounds that do not improve on the previous best by at least 1\%, or after 50 rounds. We chose these budgets to bound the cost of a run.

We ran the loop once under each condition, at the same time. The crash-allowed loop is the one analyzed in Section~\ref{sec:experiments}. The zero-crash loop is not an independent run. Its prompt contained no pointer to its own history, and in rounds 2, 3 and 6 its sessions read the history, programs or results of the crash-allowed loop. We report its round-6 weights in Table~\ref{tab:fleet} because they are the starting point of two RL baselines.

\subsection{Game Loops}
\label{sec:gameloops}

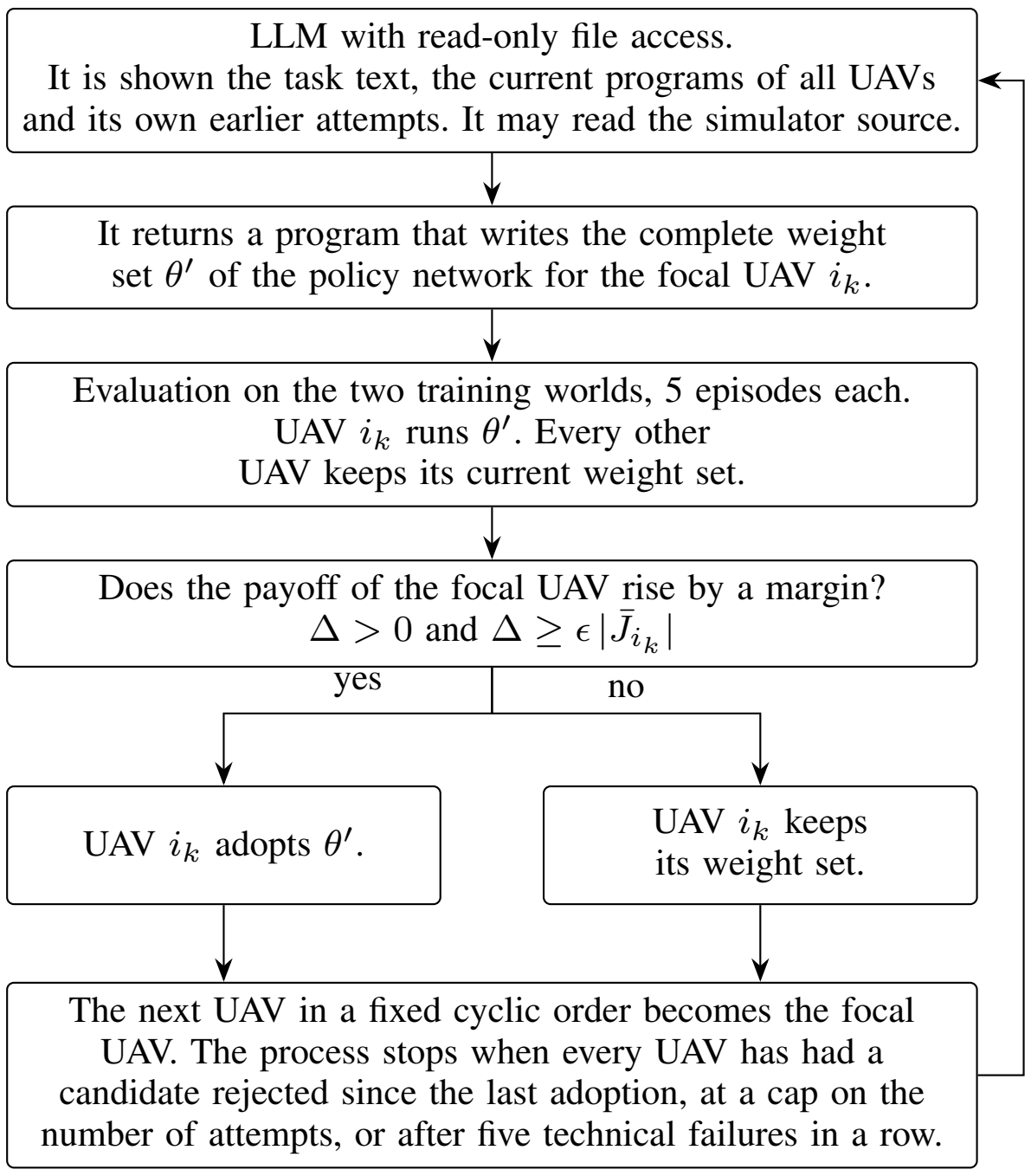
\begin{figure}[t]
\centering
\begin{tikzpicture}[font=\footnotesize, >={Stealth[length=1.8mm]},
  box/.style={draw, rounded corners=1.5pt, align=center, inner sep=3pt, text width=7.2cm},
  small/.style={draw, rounded corners=1.5pt, align=center, inner sep=3pt, text width=3.1cm, minimum height=9mm}]
\node[box] (llm) {LLM with read-only file access.\\ It is shown the task text, the current programs of all UAVs and its own earlier attempts. It may read the simulator source.};
\node[box, below=4mm of llm] (prog) {It returns a program that writes the complete weight set $\theta'$ of the policy network for the focal UAV $i_k$.};
\node[box, below=4mm of prog] (eval) {Evaluation on the two training worlds, 5 episodes each.\\ UAV $i_k$ runs $\theta'$. Every other UAV keeps its current weight set.};
\node[box, below=4mm of eval] (dec) {Does the payoff of the focal UAV rise by a margin?\\ \mbox{$\Delta > 0$ and $\Delta \ge \epsilon\,|\bar{J}_{i_k}|$}};
\node[small, below=9mm of dec, xshift=-20.5mm] (yes) {UAV $i_k$ adopts $\theta'$.};
\node[small, below=9mm of dec, xshift=20.5mm] (no) {UAV $i_k$ keeps its weight set.};
\node[box, below=24mm of dec] (next) {The next UAV in a fixed cyclic order becomes the focal UAV. The process stops when every UAV has had a candidate rejected since the last adoption, at a cap on the number of attempts, or after five technical failures in a row.};
\draw[->] (llm) -- (prog);
\draw[->] (prog) -- (eval);
\draw[->] (eval) -- (dec);
\draw[->] (dec.south) -- ++(0,-3.5mm) -| (yes.north) node[pos=0.25, above] {yes};
\draw[->] (dec.south) -- ++(0,-3.5mm) -| (no.north) node[pos=0.25, above] {no};
\draw[->] (yes.south) -- (yes.south |- next.north);
\draw[->] (no.south) -- (no.south |- next.north);
\draw[->] (next.east) -- ++(3.5mm,0) |- (llm.east);
\end{tikzpicture}
\caption{One attempt of the unilateral adoption process. $\Delta$ is the change in the focal UAV's own payoff $\bar{J}_{i_k}$, and $\epsilon$ is the adoption margin. A technical failure skips the payoff test and counts as neither an adoption nor a rejection.}
\label{fig:method}
\end{figure}

In the game loops the LLM acts for the focal UAV. The task tells it which UAV is its own, that its objective is the score of that UAV alone, and that the other UAVs run their current policies. The task shows the LLM the programs that set the current weights of the other UAVs and earlier attempts (Appendix~\ref{appx:repro}). The construction of a candidate is thus an offline step that uses the simulator source and the current policies of the other UAVs. The policy that results uses only the observation of its own UAV when it runs. The task texts of the game loops also state that every UAV belongs to a different operator, which is the assumption of this paper (Section~\ref{sec:scope}). The working directory is a reduced copy of the repository without the held-out world, the top-level test folder and the result files. Fig.~\ref{fig:method} shows one attempt, and Table~\ref{tab:loops} lists the loops.

\begin{table}[t]
\centering
\caption{Game loops of the main text. The two loops in this table use static users, the two training worlds and the margin $\epsilon = 0.01$, and they start from the round-13 program of the crash-allowed feedback loop. The 30-UAV replacement run compares the sum of the payoffs over the two worlds instead of their mean, which differs by a factor of 2 and does not change the relative test of Eq.~\eqref{eq:adopt}. Every row is one run. In the text we identify a run by its number of UAVs and its procedure.}
\label{tab:loops}
\footnotesize
\begin{tabularx}{\columnwidth}{>{\raggedright\arraybackslash}p{0.7cm}>{\raggedright\arraybackslash}p{2.2cm}>{\raggedright\arraybackslash}X}
\toprule
UAVs & Procedure & Focal UAVs and stopping rule \\
\midrule
5 & unilateral adoption, individual weight sets & UAVs 0 to 4 in order. Stops when every UAV has had a candidate rejected since the last adoption, after 100 attempts or after 5 consecutive technical failures \\
30 & imposed fleet-wide replacement, shared weight set & UAVs 0, 24, 4, 28, 17, 23, 7 and 18 in order. Set to stop after 10 consecutive attempts without an adoption or after 200 attempts. Stopped manually, with 100 completed attempts and an unfinished attempt 101 \\
\bottomrule
\end{tabularx}
\end{table}

The 5-UAV runs, including the related runs of Section~\ref{sec:games} and the replacement run of Appendix~\ref{appx:repro}, use a smaller configuration with 5 UAVs of the first type, 60 users in three clusters with the same spread, 2 channels of 1~MHz, and 3 stations with 1 charging slot and 1 waiting slot each. The stations and hotspots are those of the two training worlds. UAVs 0, 3 and 4 share a channel in the original map, and UAVs 0, 2 and 3 share a channel in the second world. Three UAVs start in the low battery band and two in the middle band. The smaller configuration lowers the cost of a run, but it also changes the UAV types, the channel groups and the capacity of the stations, so it is a separate test configuration and its scores cannot be compared with those of the 30-UAV configuration.

In the 30-UAV replacement run the 8 focal UAVs were chosen so that both UAV types, the three starting stations and the three battery bands are represented and so that the 8 UAVs use 8 different channels in each training world. The other 22 UAVs are never focal UAVs, but they run every candidate that is imposed on the fleet.

A failed LLM call, an answer without a single code block, a program that the filter rejects or that fails to apply, and a failed evaluation are technical failures. A technical failure uses up the attempt and counts neither as an adoption nor as a rejection. Five technical failures in a row stop a game loop. We ran further game loops, and Appendix~\ref{appx:repro} lists them.

\section{Experimental Setup}
\label{sec:setup}

\subsection{Evaluation Protocol}

Unless a run is identified as a single-world run, every selection and adoption decision uses the simulator of Section~\ref{sec:system} with the energy guard off, 5 episodes per world with fixed seeds, and the two training worlds. The direction mask and the battery penalty are two training mechanisms of the RL baselines (Section~\ref{sec:rl}). The evaluator disables the mask, and the evaluation configuration that we inspected sets no penalty (Appendix~\ref{appx:rl}). Scores are fleet-mean scores unless stated otherwise. Within a run the evaluation is deterministic for a fixed seed, so the evaluation of the current profile, against which each candidate is compared, is computed once and reused. Evaluations of the same weights made at other times can differ slightly (Appendix~\ref{appx:repro}).

\subsection{Reinforcement Learning Baselines}
\label{sec:rl}

For comparison we trained five MAPPO~\cite{MAPPO} configurations on the 30-UAV problem with the policy architecture of Section~\ref{sec:obs}. Table~\ref{tab:fleet} lists them. They involve four mechanisms. The energy guard is the rule of Section~\ref{sec:modes}. The \emph{direction mask} is a training-time restriction that narrows the allowed directions of horizontal movement to a sector around the direction of a station as the battery falls from $1.5R$ to $R$. The masking of the charge logit (Section~\ref{sec:obs}) is part of the environment in every run. The \emph{battery penalty} subtracts 150 from the reward of a flying UAV in a step in which its battery fraction is below a threshold $\tau$ and either the UAV is farther than 100~m from its nearest station and does not move closer to it, or charging is selectable and the UAV does not choose charge. We set $\tau$ to 0.14 and 0.35 to match the approximate battery fractions at which the two LLM-written programs of Section~\ref{sec:program} switch to a return to a station.

According to our run records, the two warm-start configurations begin from the round-6 weights of the zero-crash loop. They first train only the critic for 50 updates, which count toward the stated number of steps. Appendix~\ref{appx:rl} gives the main hyperparameters and the exact rules.

\subsection{Scope of the Evidence}
\label{sec:scope-evidence}

Every configuration in this paper was run once, and every policy was evaluated with the same 5 episode seeds per world. The reported differences are therefore single observations and not estimates with a known uncertainty. The programmed adoption tests and the ranking of rounds used the scores on the two training worlds only. The two penalty configurations and the warm start trained without the guard were specified after the first evaluation batch, which included the held-out world, was available. The set of baselines was therefore not fixed in advance. Our planning notes for these configurations quote only training-world results, which does not rule out an influence of the held-out results. Section~\ref{sec:threats} discusses these and further limits.

\section{Results}
\label{sec:experiments}

\subsection{Unilateral Adoption and Imposed Fleet-Wide Replacement}
\label{sec:games}

Table~\ref{tab:game} summarizes the outcomes of the game loops of Table~\ref{tab:loops}. All of them start from the round-13 program of the crash-allowed feedback loop, whose results Sections~\ref{sec:fleet} to~\ref{sec:rounds} report. All fleet means in this subsection are measured on the worlds and initial conditions that the adoption tests use.

\begin{table*}[t]
\centering
\caption{Outcomes of the game loops. Fleet means average the scores over the two training worlds. The row for unilateral adoption gives the starting and the final mean. The row for replacement gives the starting mean and the recorded range from attempt 2 through the last fleet evaluation. The fourth column states whether a run ended because candidates stopped being adopted. The replacement run was stopped manually.}
\label{tab:game}
\footnotesize
\begin{tabularx}{\textwidth}{>{\raggedright\arraybackslash}X>{\raggedright\arraybackslash}p{1.3cm}>{\raggedright\arraybackslash}p{1.4cm}>{\raggedright\arraybackslash}p{3.7cm}>{\raggedright\arraybackslash}p{2.8cm}}
\toprule
Run & Attempts & Adoptions & Ended by rejections & Fleet mean \\
\midrule
5 UAVs, unilateral adoption & 70 & 23 & yes, one full cycle without an adoption & 84.7 to 105.9 \\
30 UAVs, fleet-wide replacement & 100 & 79 & no, 3 programs repeated in a cycle of 8 attempts & 149.1 to 131.4--133.4 \\
\bottomrule
\end{tabularx}
\end{table*}

\subsubsection{Unilateral adoption with 5 UAVs}

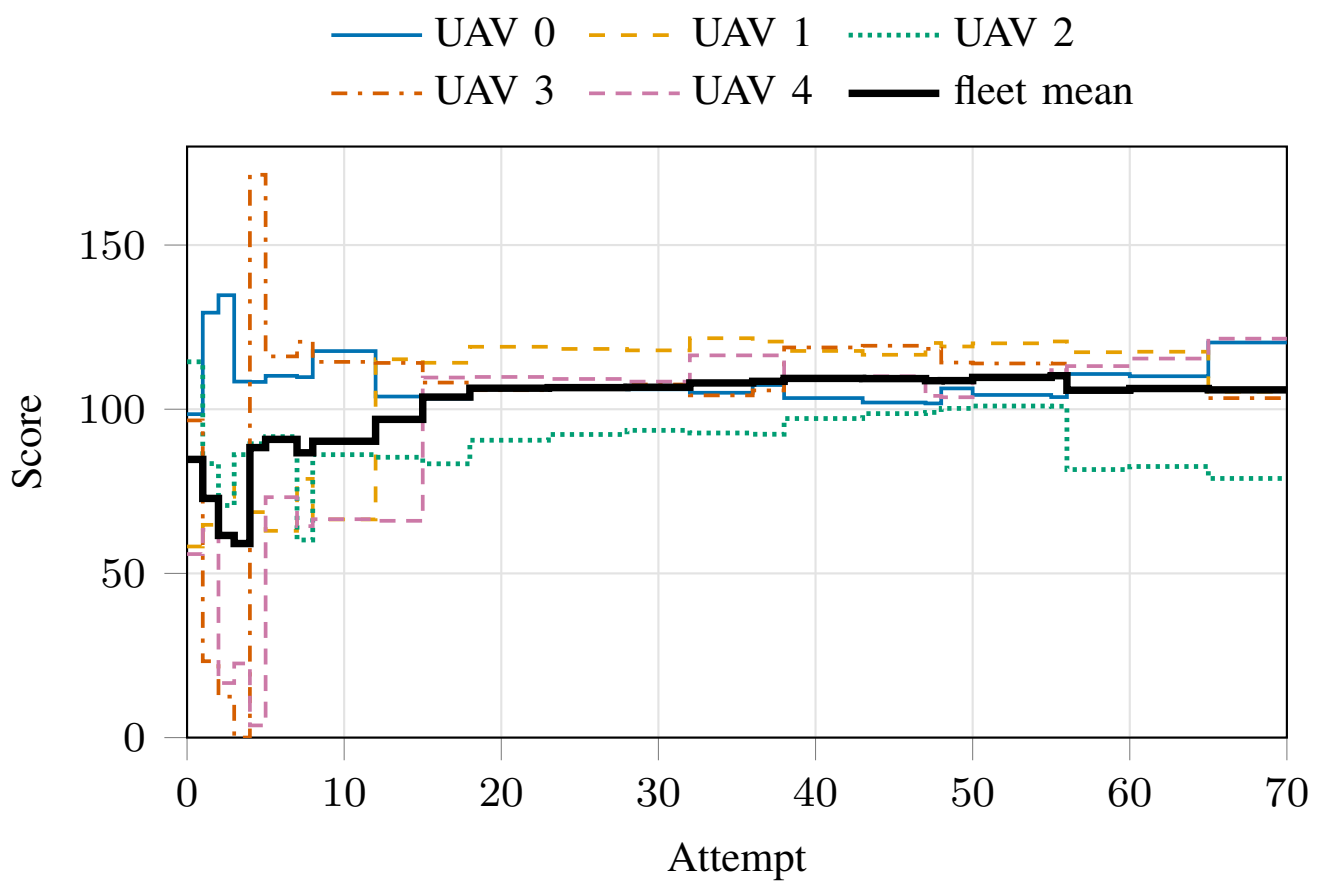
\begin{figure}[t]
\centering
\begin{tikzpicture}
\begin{axis}[paper, height=0.62\columnwidth, xlabel={Attempt}, ylabel={Score},
  xmin=0, xmax=70, ymin=0, ymax=180, legend columns=3,
  legend style={at={(0.5,1.03)}, anchor=south, /tikz/every even column/.append style={column sep=4pt}}]
\addplot[const plot, cBlue, line width=0.7pt] table[x=attempt, y=u0]{figs/data/unilateral-5uav.dat}; \addlegendentry{UAV 0}
\addplot[const plot, cOrange, dashed, line width=0.7pt] table[x=attempt, y=u1]{figs/data/unilateral-5uav.dat}; \addlegendentry{UAV 1}
\addplot[const plot, cGreen, densely dotted, line width=0.9pt] table[x=attempt, y=u2]{figs/data/unilateral-5uav.dat}; \addlegendentry{UAV 2}
\addplot[const plot, cRed, dash dot, line width=0.7pt] table[x=attempt, y=u3]{figs/data/unilateral-5uav.dat}; \addlegendentry{UAV 3}
\addplot[const plot, cPurple, densely dashed, line width=0.7pt] table[x=attempt, y=u4]{figs/data/unilateral-5uav.dat}; \addlegendentry{UAV 4}
\addplot[const plot, black, line width=1.4pt] table[x=attempt, y=fleet]{figs/data/unilateral-5uav.dat}; \addlegendentry{fleet mean}
\end{axis}
\end{tikzpicture}
\caption{Unilateral adoption with 5 UAVs and $\epsilon = 0.01$: score of each UAV and fleet mean under the profile in force after each attempt. The run stopped after 70 attempts and 23 adoptions.}
\label{fig:unilateral}
\end{figure}

The 5-UAV unilateral run stopped by its own rule after 70 attempts and 23 adoptions, and the fleet mean ended 25\% above its start (Fig.~\ref{fig:unilateral}). The fleet mean did not rise monotonically, because an adoption that raised the score of one UAV could lower the scores of others. The first adoption, by UAV~0, raised its score from 98.5 to 129.4. It lowered the score of UAV~2 from 114.4 to 83.4 and the score of UAV~3 from 96.6 to 23.3, and it lowered the fleet mean from 84.7 to 72.8. UAV~3, which lost most, shares a channel with UAV~0 on both training worlds, and UAV~2 shares one with UAV~0 on the second world. UAV~4 shares a channel with UAV~0 only on the original map, and its score rose from 55.9 to 63.3. This pattern is consistent with same-channel interference, but we did not measure the cause. The UAVs that lost then made their own attempts. After the first three adoptions the fleet mean was 59.1 and the score of UAV~3 was close to zero. The fourth adoption was made by UAV~3 itself, raised its score to 171.4 and brought the fleet mean to 88.3. The candidates adopted in attempts 2 to 5 matched the first one. Apart from comments, the programs that UAVs~1 to 4 adopted were the program that UAV~0 had adopted in attempt 1, which the task showed to the LLM as the current program of UAV~0. After attempt 5 all five UAVs ran the same program, and the fleet mean was 90.8. Later adoptions changed the programs of individual UAVs, except that of UAV~3, which made no further adoption. They also lowered the scores of UAVs that kept their weights. In attempt 56, for example, the adoption by UAV~0 lowered the score of UAV~2 from 100.9 to 81.6, and the next three candidates of UAV~2 changed its payoff by $-3.55\%$, $+0.03\%$ and $-1.30\%$, so none of them was adopted. At the end the scores of UAVs~0 to 4 were 120.3, 105.4, 78.9, 103.4 and 121.5. Four UAVs ended above their starting scores, and UAV~2 ended below its starting score of 114.4. In the final cycle the five candidates changed the payoffs of their focal UAVs by $-3.83\%$, $+0.69\%$, $-1.30\%$, $+0.30\%$ and $-40.22\%$. Two of them were improvements below the margin, so the stop does not mean that no improving weight set exists.

Two related runs with 5 UAVs also used unilateral adoption. The first started from the same profile as the 5-UAV unilateral run above and used the original map alone. It stopped after 20 attempts and 9 adoptions, and its fleet mean on that map rose from 84.8 to 107.1. The second started from the final profile of the first and used both training worlds. None of its five candidates was adopted, and the fleet mean of that profile on the two worlds was 106.3, with scores of 113.2, 112.2, 99.2, 103.3 and 103.6. The 5-UAV unilateral run above and this continuation thus ended at different profiles, with fleet means of 105.9 and 106.3 on the same two worlds and different scores for the individual UAVs.

\subsubsection{Imposed fleet-wide replacement with 30 UAVs}

\begin{figure}[t]
\centering
\begin{tikzpicture}
\begin{axis}[paper, width=0.98\columnwidth, height=0.48\columnwidth, xlabel={Attempt}, ylabel={Fleet mean},
  xmin=0, xmax=100, ymin=128, ymax=151]
\addplot[const plot, black, mark=*, mark size=0.7pt] table[x=round, y=fleet]{figs/data/replacement-30uav.dat};
\end{axis}
\end{tikzpicture}
\caption{Imposed fleet-wide replacement with 30 UAVs, static users and $\epsilon = 0.01$: fleet mean over the two training worlds under the policy in force after each attempt. The run did not meet its stopping rule and was stopped manually. The last plotted point is attempt 98, because the record holds no fleet evaluation of the policy adopted in attempt 100.}
\label{fig:fleet}
\end{figure}
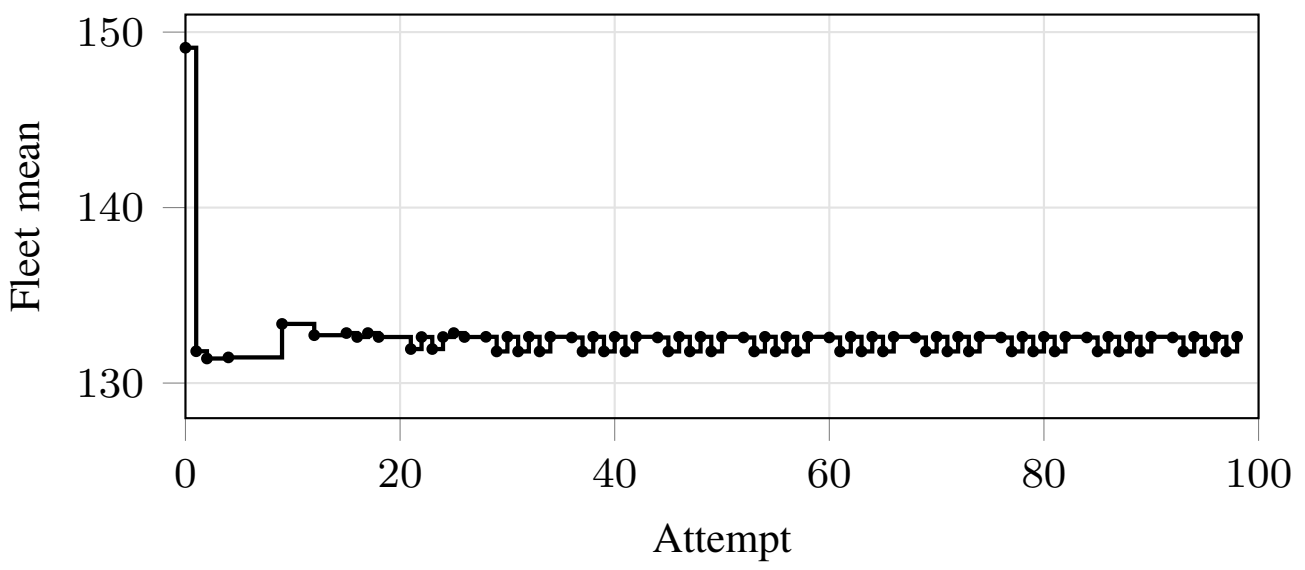

The 30-UAV replacement run never met its stopping rule (Fig.~\ref{fig:fleet}). In this run the first candidate raised the payoff of its focal UAV by 15.5\%, and the fleet mean fell from 149.1 to 131.8 once all 30 UAVs ran it. In 100 attempts, 79 candidates passed the adoption test, and the fleet mean stayed between 131.4 and 133.4. The fleet evaluations of this run end at attempt 98. The candidate of attempt 100 was adopted, but the record holds no fleet evaluation of the resulting policy. From attempt 37 on, the program adopted in an attempt was identical, apart from comments, to the program adopted 8 attempts earlier, and every candidate for one of the 8 focal UAVs, UAV~4, was rejected. Only three distinct programs occurred in this cycle, and the fleet mean took the three values 131.80, 132.60 and 132.64. We compared the program texts and did not compare the resulting weights. No UAV crashed, and all 360 users stayed connected. Between the starting profile and the profile evaluated at attempt 98, the mean number of transmitting UAVs on the original map rose from 7.3 to 15.6, and the mean spectral efficiency per user fell from 12.43 to 10.98~bit/s/Hz.

Seven further 30-UAV runs with static users and imposed fleet-wide replacement (Appendix~\ref{appx:repro}), which started from other programs or used a single training world, never had 10 consecutive rejections. Their final fleet means lay between 40\% below and 1.6\% above their starting values. Appendix~\ref{appx:repro} also reports a run with 5 UAVs and imposed fleet-wide replacement, which compares the two procedures at the same fleet size.

\subsection{LLM-Written Weights and MAPPO}
\label{sec:fleet}

Table~\ref{tab:fleet} compares five LLM-written weight sets with the five MAPPO configurations in the single-fleet setting. The first row is the self-interested weight set of Section~\ref{sec:oneshot}. Its task counts only the own reward of each UAV. The MAPPO configurations are also trained on the own reward of each UAV, two of them with an added battery penalty (Appendix~\ref{appx:rl}), so this row is the closest match of objectives. The other four LLM rows were written for the fleet-mean score. We make three observations.

First, without the energy guard the self-interested weight set scores 133.7 and 135.9 on the training worlds and loses 0 and 2.0 UAVs per episode. On the held-out world it scores 113.9 and loses 3.6. The four weight sets that were written for the fleet-mean score crash no UAV on any of the three worlds. Every MAPPO configuration loses 22.6 to 30 of its 30 UAVs per episode on the training worlds. The crashes also occur in the three configurations that were trained without the guard, two of which used the battery penalty.

Second, with the guard on, the MAPPO configurations score 131.1 to 136.4 on the two training worlds, and the self-interested weight set scores 133.3 and 135.5. The one-shot weights for the fleet-mean score reach 126.4 to 145.1, and the round-13 weights of the crash-allowed loop score 149.2 and 148.7. On the held-out world the gap between the LLM-written weights and the MAPPO configurations is smaller. The four weight sets for the fleet-mean score reach 100.7 to 116.9 without a crash, the self-interested weight set scores 113.8 with 1.8 crashed UAVs per episode, and the MAPPO configurations score 86.6 to 114.3 with 0.2 to 3.0 crashed UAVs per episode.

Third, to see whether a policy enters charge mode on its own or only when the guard forces it, we compared two counts in the evaluations with the guard on. Let $C$ be the number of entries into charge mode across the fleet per episode, and let $G$ be the corresponding number of guard engagements. An entry into charge mode does not establish that charging began. For the policy trained from scratch, the logs on the original map give $C = 96.0$ and $G = 96.0$. On the original map, the relative difference $(C - G)/C$ is about 9\% for the warm start trained with the guard and 16\% to 53\% for the three configurations trained without the guard. These are differences of counts per episode. Guard engagements were not matched to individual entries into charge mode, so the ratios do not identify entries that a policy chose on its own. The four weight sets for the fleet-mean score never engage the guard on the original map or on the held-out world. On the second world the guard engages 1.2 times per episode for three of the four weight sets and 66.8 times for the round-13 weights, whose score there is 148.7 with the guard and 149.1 without it. For the self-interested weight set the guard engages 73.6, 87.2 and 24.0 times per episode on the original map, the second world and the held-out world. These counts show that three of the four weight sets for the fleet-mean score almost never needed the guard. They do not show how often a MAPPO policy entered charge mode on its own.

The fleet-mean score is the social welfare per UAV (Section~\ref{sec:problem}), so Table~\ref{tab:fleet} also shows what crashes cost the fleet. The MAPPO configurations that lose 22.6 to 30 UAVs per episode without the guard score 26.2 to 113.1 on the training worlds, against 131.1 to 136.4 with the guard. The few crashes of the self-interested weight set did not lower its fleet mean in the same way. On the second world it scores 135.9 with 2.0 crashes per episode without the guard and 135.5 with no crash with the guard. The 6 UAVs that crashed in at least one of the 5 episodes average 154.3 without the guard and 161.1 with it, and the other 24 UAVs average 131.3 and 129.1. We did not examine why these 6 UAVs score higher. On the held-out world the pattern is reversed. There the 12 UAVs that crashed average 52.0 without the guard, and the other 18 UAVs average 155.1. The guard also changes the behavior of UAVs that do not crash, so this comparison does not isolate the effect of the crashes.

\begin{table*}[t]
\centering
\caption{Single-fleet setting with 30 UAVs: each cell gives the fleet-mean score / mean number of crashed UAVs per episode, with the energy guard on and off at evaluation time. The two training worlds are the original map and the second world. Each MAPPO row names the training budget and the mechanisms that were active in training. A dash marks an evaluation that was not run.}
\label{tab:fleet}
\footnotesize
\begin{tabularx}{\textwidth}{>{\raggedright\arraybackslash}Xcccccc}
\toprule
 & \multicolumn{3}{c}{Guard on} & \multicolumn{3}{c}{Guard off} \\
\cmidrule(lr){2-4}\cmidrule(lr){5-7}
Policy & Original & Second & Held-out & Original & Second & Held-out \\
\midrule
LLM, one-shot, self-interested & 133.3 / 0 & 135.5 / 0 & 113.8 / 1.8 & 133.7 / 0 & 135.9 / 2.0 & 113.9 / 3.6 \\
LLM, one-shot, crash-allowed & 145.1 / 0 & 130.2 / 0 & 116.1 / 0 & 145.1 / 0 & 130.2 / 0 & 116.1 / 0 \\
LLM, one-shot, zero-crash & 128.5 / 0 & 126.4 / 0 & 100.7 / 0 & 128.5 / 0 & 126.4 / 0 & 100.7 / 0 \\
LLM, crash-allowed loop, round 13 & 149.2 / 0 & 148.7 / 0 & 116.9 / 0 & 149.2 / 0 & 149.1 / 0 & 116.9 / 0 \\
LLM, zero-crash loop, round 6 & 144.1 / 0 & 145.8 / 0 & 116.7 / 0 & 144.1 / 0 & 145.8 / 0 & 116.7 / 0 \\
\midrule
MAPPO from scratch, 20~M steps, guard and direction mask & 131.1 / 0 & 136.1 / 0 & 114.3 / 1.8 & 26.2 / 30 & 27.0 / 30 & 23.5 / 30 \\
MAPPO warm start, 10~M steps, guard and direction mask & 132.8 / 0 & 135.0 / 0 & 112.4 / 3.0 & 43.4 / 30 & 47.7 / 30 & 42.5 / 30 \\
MAPPO, battery penalty $\tau = 0.14$, 10~M steps, no guard & 133.4 / 0.2 & 135.0 / 0 & 86.6 / 1.8 & 113.1 / 22.6 & 46.4 / 30 & -- \\
MAPPO, battery penalty $\tau = 0.35$, 10~M steps, no guard & 136.4 / 0 & 136.3 / 0 & 110.2 / 0.2 & 45.7 / 30 & 47.9 / 30 & -- \\
MAPPO warm start, 10~M steps, no guard & 132.6 / 0 & 132.9 / 0 & 112.2 / 1.0 & 70.8 / 30 & 48.5 / 30 & -- \\
\bottomrule
\end{tabularx}
\end{table*}

\subsection{What the Written Programs Do}
\label{sec:program}

The programs that the LLM returned state their rules in the weights of the first hidden layer. In the round-13 program of the crash-allowed loop, when the battery input falls below about 14\%, a hidden unit switches the movement head from seeking users to a return vector toward the station that the UAV last left. A second unit that is driven by the same battery signal raises the charge logit, so the UAV chooses charge as soon as charging becomes selectable. The round-6 program of the zero-crash loop uses a threshold of about 35\%.

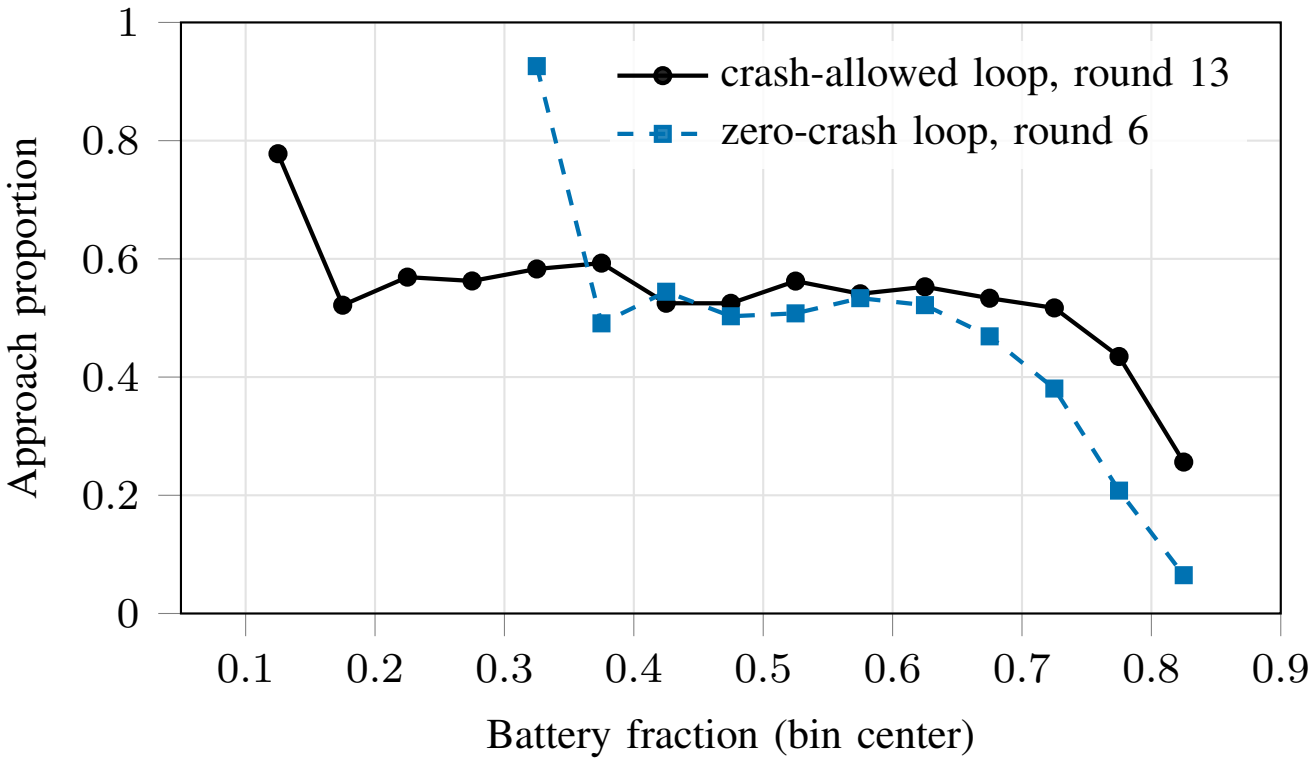
\begin{figure}[t]
\centering
\begin{tikzpicture}
\begin{axis}[paper, xlabel={Battery fraction (bin center)}, ylabel={Approach proportion},
  xmin=0.05, xmax=0.9, ymin=0, ymax=1.0, xtick={0.1,0.2,...,0.9}, legend pos=north east]
\addplot[black, mark=*, mark size=1.4pt] table[x=b, y=p]{figs/data/approach-round13.dat};
\addlegendentry{crash-allowed loop, round 13}
\addplot[cBlue, dashed, mark=square*, mark size=1.3pt, mark options={solid}] table[x=b, y=p]{figs/data/approach-round6.dat};
\addlegendentry{zero-crash loop, round 6}
\end{axis}
\end{tikzpicture}
\caption{Proportion of steps in which a flying UAV that is more than 100~m from its nearest station moves closer to that station, by battery fraction in bins of width 0.05. The steps are pooled from 15 recorded episodes per program, 5 on each of the three worlds, with the energy guard off. The steps of an episode are not independent observations. Bins with fewer than 1,000 steps are omitted.}
\label{fig:approach}
\end{figure}

As a diagnostic after the runs, we compared these thresholds with recorded episodes of the two programs on the three worlds with the guard off (Fig.~\ref{fig:approach}). For every step of a flying UAV that is more than 100~m from its nearest station, we recorded whether the UAV moved closer to that station, and we grouped the steps by battery fraction. The proportion of approaching steps drops from 0.78 to 0.52 between the bins below and above a battery fraction of 0.15 for the round-13 program, and from 0.93 to 0.49 around 0.35 for the round-6 program. The bins measure the approach to the nearest station and not the return vector itself, and their width limits the precision to 0.05. We located each drop as the largest fall between neighboring bins below a battery fraction of 0.6 that both hold at least 1,000 steps. Section~\ref{sec:threats} notes how we chose this criterion.

In the reported deterministic evaluations without the guard, none of the evaluated MAPPO policies kept the whole fleet alive (Table~\ref{tab:fleet}). This includes the two warm starts. According to our run records they began from the round-6 weights of the zero-crash loop, which crash no UAV on any of the three worlds, and after 10~M steps of training they lost all 30 UAVs per episode on both training worlds without the guard. We have not examined why the training removed this property, and we did not apply the measurement of Fig.~\ref{fig:approach} to the MAPPO policies. The comparison is also not controlled for the information available to each method. The LLM could read the energy and charging logic in the simulator source, while the RL agents had to learn it from observations and rewards.

\subsection{Feedback Loop and Ablations}
\label{sec:rounds}

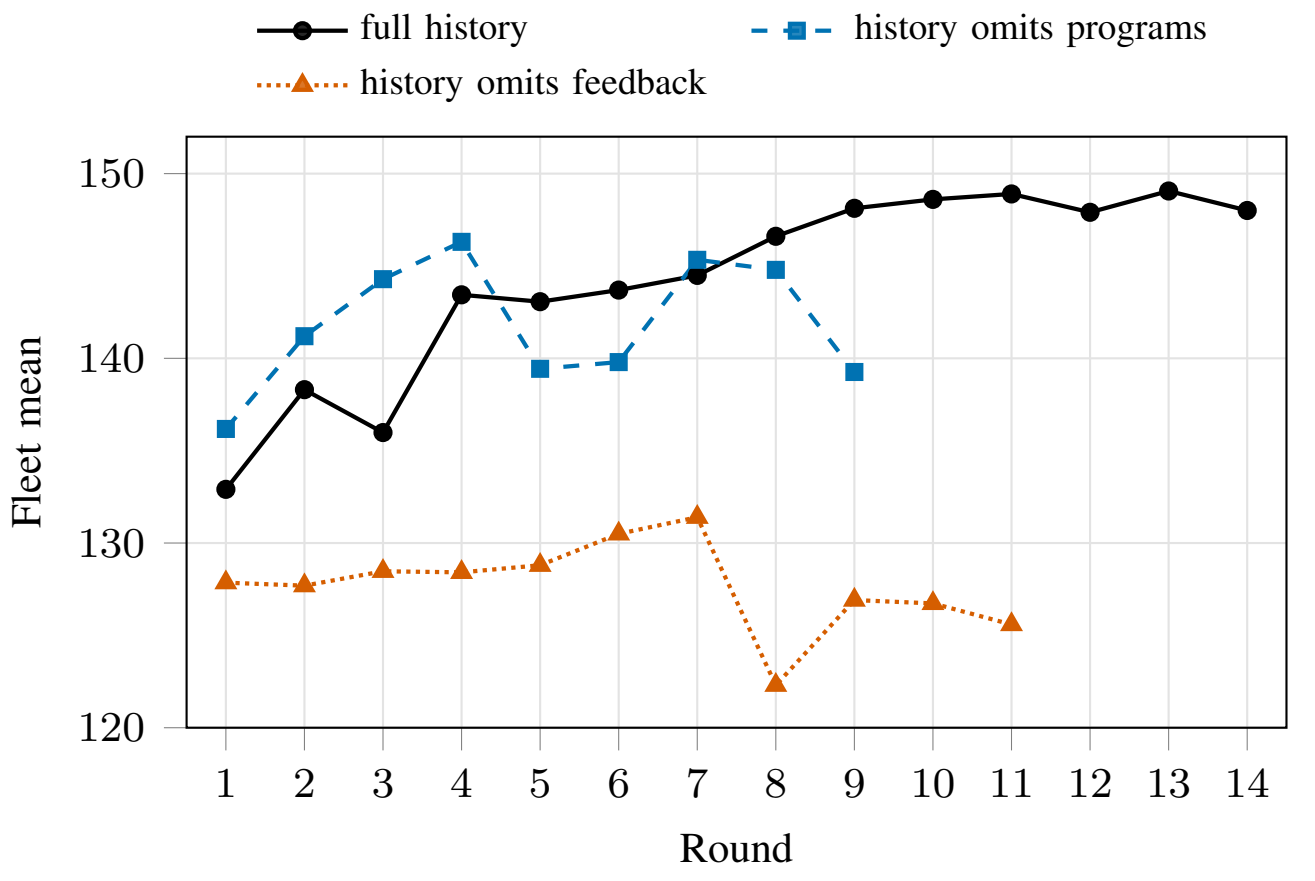
\begin{figure}[t]
\centering
\begin{tikzpicture}
\begin{axis}[paper, xlabel={Round}, ylabel={Fleet mean},
  xmin=0.5, xmax=14.5, ymin=120, ymax=152, xtick={1,2,...,14},
  legend columns=2, legend style={font=\scriptsize, at={(0.5,1.03)}, anchor=south, /tikz/every even column/.append style={column sep=6pt}}]
\addplot[black, mark=*, mark size=1.4pt] table[x=round, y=score]{figs/data/loop-full-history.dat};
\addlegendentry{full history}
\addplot[cBlue, dashed, mark=square*, mark size=1.3pt, mark options={solid}] table[x=round, y=score]{figs/data/loop-without-programs.dat};
\addlegendentry{history omits programs}
\addplot[cRed, densely dotted, mark=triangle*, mark size=1.7pt, mark options={solid}] table[x=round, y=score]{figs/data/loop-without-feedback.dat};
\addlegendentry{history omits feedback}
\end{axis}
\end{tikzpicture}
\caption{Crash-allowed feedback loop with 30 UAVs: mean of the fleet-mean score over the two training worlds in each round, for the full history, for a history without the earlier programs and explanations, and for a history without the evaluator's feedback. Each history variant was run once.}
\label{fig:loop}
\end{figure}

Fig.~\ref{fig:loop} shows the crash-allowed feedback loop. Its fleet mean rose from 132.9 in round 1 to 149.1 in round 13, the round with the highest score, which we selected. The loop stopped after round 14, because rounds 10 to 14 did not improve on the best score by at least 1\%. A single answer under the same condition scored between 127.9 and 137.9 in the runs' own evaluations. The lower values are round 1 of this loop and of the two runs below, and 137.9 is the run's own evaluation of the crash-allowed one-shot answer. Table~\ref{tab:fleet} instead gives evaluations made after the runs (Appendix~\ref{appx:repro}).

Two further runs changed the history that the loop returns to the LLM. One kept the evaluator's scores, metrics and episode summaries but not the earlier programs and explanations, and it reached 146.3 in round 4 and stopped after round 9. The other kept the earlier programs and explanations but not the evaluator's feedback, and it stayed between 122.3 and 131.4 and stopped after round 11. Both runs used the crash-allowed condition and had the reference weight set. In the run without the earlier programs and explanations, round 4 lost 2.4 UAVs per episode on the second world, and round 7 lost 12 UAVs per episode on each world. In the run without the evaluator's feedback, every round lost 12 UAVs per episode on each world. The prompts of the three runs already differed in round 1, where they scored 132.9, 136.2 and 127.9, and each run was made once, so these runs do not show which part of the history the full loop relies on.

A further pair of one-shot runs removed the reference weight set, the section of the task that describes it and the other mentions of it in the task, and it used a reduced working directory (Appendix~\ref{appx:repro}). These changes are not separated. In the zero-crash condition the returned program kept all 30 UAVs alive and scored 123.8 and 139.9 on the two training worlds. In the crash-allowed condition the returned program scored 141.6 and 129.7, and the mean crash count was 12 of 30 UAVs per episode on each world. Its explanation states that the 12 UAVs of the second type ``remain silent and never charge.'' This choice removes their interference and their demand for charging. In this single run, the crash-allowed condition without the reference weight set did not produce crash-free weights.

\section{Discussion}
\label{sec:threats}

\subsection{What the Observations Suggest}

In these runs, the gain of a focal UAV against a fixed fleet did not ensure a higher fleet mean. In the 5-UAV unilateral run the first three adoptions raised the scores of the adopting UAVs and lowered the fleet mean. Once all five UAVs ran the program that UAV~0 had adopted first, the fleet mean was 90.8, which is above its starting value of 84.7, and the score of UAV~0 was 110.2 instead of 129.4. In the 30-UAV replacement run, the first replacement lowered the fleet mean from 149.1 to 131.8. We suspect that part of such a gain is obtained at the expense of the other UAVs, for example by taking their users or by transmitting when they do not, and that this part disappears when all UAVs behave in the same way. The rise in the number of transmitting UAVs and the fall in spectral efficiency in the 30-UAV replacement run are consistent with this hypothesis, but they do not identify the cause. In the 5-UAV unilateral run the first adoption hurt most the UAV that shares a channel with UAV~0 on both training worlds, which is also consistent with this hypothesis and was also not measured (Section~\ref{sec:games}). We have not measured how much of a loss comes from interference, from user handovers or from competition for charging slots.

In attempts 2 to 5 of the 5-UAV unilateral run the LLM returned, for UAVs~1 to 4, the program of UAV~0, which the task showed to it. The task exposed the program that the LLM later returned, and we did not isolate the effect of this exposure on its output.

The runs also differed in how they ended. The 5-UAV unilateral run reached a profile at which five tested candidates in a row were rejected, and the 30-UAV replacement run repeated the same programs in cycles. The outcome of the 5-UAV unilateral run is one observed case.

\subsection{Threats to Validity}

\emph{Single runs.} Every configuration was run once. Three first-round calls whose prompts differ in one paragraph scored 127.9 to 136.2 (Section~\ref{sec:rounds}), while the first candidates of the 5-UAV unilateral run and the 5-UAV replacement run of Appendix~\ref{appx:repro} were the same program apart from the position of two statements, and we did not repeat any loop. The adoption margin of 1\% was not compared with this variation. The observation that unilateral adoption stopped at a higher fleet mean rests on one run with 5 UAVs and on one further run on a single world from the same starting profile, whose final profile then went through five attempts on both worlds without an adoption. In both runs the focal UAVs were visited in a fixed order that starts with UAV~0, and we did not test whether another order changes which UAVs gain. The two runs ended at different profiles (Section~\ref{sec:games}). They differ both in the sequence of adoptions and in the worlds that they used, so they do not show which of the two determines where the process stops.

\emph{Scope.} All results that this paper analyzes are for static users, and unilateral adoption was run to its stopping rule only with 5 UAVs. The main text compares a 5-UAV run with unilateral adoption and a 30-UAV run with imposed fleet-wide replacement, so the difference between the two procedures there is mixed with the difference in fleet size. Appendix~\ref{appx:repro} gives the comparison at the same fleet size. We also ran four runs with moving users, which this paper does not analyze (Appendix~\ref{appx:repro}). The outcome of the 5-UAV unilateral run should therefore not be assumed for larger fleets or for moving users.

\emph{Comparison with RL.} The set of RL baselines was extended after the first evaluation results were available. The baselines were trained for 10 or 20~M steps with settings that we did not tune. Two of the configurations started from crash-free LLM-written weights, and after training they lost all 30 UAVs per episode on both training worlds without the guard. We have not examined why their training removed this property (Section~\ref{sec:program}). The LLM could read the simulator source, which the RL agents could not. The comparison therefore shows what these configurations reached and not what RL can reach.

\emph{Isolation of the LLM.} The read-only sandbox did not confine reads to the working directory. In the single-fleet runs the sessions read files outside it, and the zero-crash loop read material of the crash-allowed loop (Appendix~\ref{appx:repro}). The name of the held-out world file appears in file listings of nine sessions, although no session printed its content. In the ablations of Section~\ref{sec:rounds} the LLM could still reach the directories of earlier runs.

\emph{Evaluation.} All scores come from 5 fixed initializations per world. The round selected from the feedback loop was chosen on the two training worlds, and its advantage over the other policies of Table~\ref{tab:fleet} is smaller on the held-out world. The 5-UAV world differs from the 30-UAV world in the number of users, UAV types, channels and station capacity. The criterion used in Fig.~\ref{fig:approach} to locate the drop was chosen after a first criterion had failed for one of the two programs.

\emph{Interpretation of the game.} A stop of the adoption process says something about the tested candidates and nothing about equilibrium (Section~\ref{sec:adoption}). We assume that the players are independent operators, and we do not establish how the score relates to the benefit of an operator or who arranges the shared channels and stations (Section~\ref{sec:scope}).

\emph{Availability.} The simulator, the candidate programs and the run logs are not part of this preprint (Appendix~\ref{appx:repro}).

\emph{Related work.} The statement that we found no source in which several self-interested players of a game adopt LLM-written network weights in turn rests on the search of Appendix~\ref{appx:search}. New work on this topic appears quickly, and the search should be repeated.

\section{Conclusion}
\label{sec:conclusion}

We let an LLM improve the UAVs of a game one at a time. In each attempt the LLM writes a program that sets every weight of the policy of one UAV, and the UAV adopts the weights only when its own score rises by a margin while the others keep theirs. In one run with 5 UAVs, this process stopped after 70 attempts and left the fleet mean 25\% higher, although early adoptions had first lowered it and one UAV ended below its starting score. In one run with 30 UAVs in which each adopted candidate was instead imposed on the whole fleet, the process never met its stopping rule and repeated the same programs in cycles, with a fleet mean below its start.

Without an energy guard, the four weight sets that the LLM wrote for the fleet-mean score of 30 UAVs lost no UAV on any of the three worlds, and the weights for a task in which each UAV counts only its own reward lost 0 and 2.0 UAVs per episode on the training worlds and 3.6 on the held-out world. When it wrote the four weight sets for the fleet-mean score, the LLM had a reference weight set, and for two of them also a task condition that forbids crashes. UAVs crashed in a one-shot run that had neither and in some or all rounds of two loops that returned only part of the history, although the LLM had the reference weight set in these loops. Each of the five MAPPO configurations that we trained lost at least 22.6 of its 30 UAVs per episode on the training worlds with the guard off. This comparison is limited to these five configurations, and the LLM could read the energy and charging logic that the RL agents had to learn. In unilateral adoption an attempt costs one LLM answer and 10 evaluation episodes, and the adopted weights run without LLM calls.

Each of these statements rests on single runs with static users. Several questions remain open. The first is how the size of the margin affects whether and where the process stops. The second is whether a profile at which the process stops is an $\epsilon$-equilibrium. The third is how the gain of a focal UAV against fixed opponents relates to the later scores of that UAV and of the fleet. The fourth is whether another order of the focal UAVs changes which UAVs gain. The last is whether the outcome of the 5-UAV unilateral run holds for larger fleets and for moving users.

\appendices
\section{Simulator Details}
\label{appx:sim}

This appendix adds details to Section~\ref{sec:system}. All statements were checked against the simulator source code.

\subsection{Energy}

Table~\ref{tab:energy} gives the per-step energies, which are Yang's power constants multiplied by the step length of 10~s. A UAV that flies to a station in charge mode pays the larger of its movement energy and the hover energy, plus the descent energy from its altitude on arrival. In charge mode, a UAV arrives when its horizontal distance to the target station after the move is at most 100~m. It is then placed at the station without further horizontal flight energy. The descent energy is deducted before the station assigns a slot, and it is not refunded if the station rejects the UAV. The simulator checks for a crash after it deducts the flight and descent energy and before it assigns the slots. A rejection also costs one step of hover energy and is followed by another crash check. Before the last step of an episode, a rejected UAV that has not crashed stays above the station at its previous flight altitude. The energy updates at a station use the slot roles from before the admission of that step, so a new arrival gains no charging energy and pays no idle energy in the step of its arrival. A UAV in a waiting slot draws the idle energy. On the step in which a UAV departs from a station, it pays the climb energy for 50~m and is placed above the station cell at 50~m. A UAV in a charging slot gains the base charging energy multiplied by Yang's state-of-charge factor. For a state of charge $s$ this factor is $0.841 + 0.159\,s/0.8$ if $s \le 0.8$ and $(1 - s)/0.2$ otherwise. The implementation reads the factor from a table of 512 points with index $\lfloor 511\,s \rfloor$. It first predicts the state of charge after the step from the current value, and it then computes the energy gain with the factor at the midpoint of the current and the predicted value.

\begin{table}[t]
\centering
\caption{Movement energies, per-step energies and battery capacity, in joules, for the two UAV types (Yang's constants).}
\label{tab:energy}
\footnotesize
\begin{tabular}{lrr}
\toprule
Quantity & First type & Second type \\
\midrule
Flight, per 200~m of horizontal travel & 2,856.8 & 1,676.7 \\
Climb, per 50~m & 4,353.9 & 1,908.6 \\
Descent, per 50~m & 1,658.9 & 561.1 \\
Service & 3,186.4 & 1,354.8 \\
Hover & 3,006.4 & 1,234.8 \\
Idle in a waiting slot & 90 & 37 \\
Base charging energy & 4,142.3 & 1,761.2 \\
Battery capacity & 735,400 & 367,700 \\
\bottomrule
\end{tabular}
\end{table}

\subsection{Charging Stations}

At each station and in each step, a safety rule first selects at most one waiting UAV. It is the waiting UAV whose battery would last the fewest whole steps at the idle draw, if that number is below 10. This UAV takes a free charging slot, and if no charging slot is free it displaces the charging UAV with the smallest estimate $h = \max\{1, \lceil (E_{\max} s_d - E)^{+} / \max(B\,f((s + s_d)/2), 10^{-6}) \rceil\}$ of the steps to its departure. Here $E$ is the battery energy of that UAV, $E_{\max}$ its capacity, $B$ its base charging energy per step, $s$ its state of charge, $s_d$ the state of charge at which it departs, $f$ the charging factor from the formula above and not from the table, and $(x)^{+} = \max(x, 0)$. Equal estimates favor the first charging slot. The priority rule of Section~\ref{sec:energy} then fills the remaining free charging slots. A UAV in a waiting slot leaves the station only when it is forced out, which happens when its state of charge is at least 0.8 or when it has been displaced from a charging slot. If every station has rejected a UAV, its choice of a target station ignores the rejections. These rules are Yang's, except for the handling of rejections and the straight flight to the station, which are ours.

\subsection{Episode Seeds}

Episode $m \in \{0, \dots, 4\}$ of an evaluation uses the seed $(17 + 1000000007\,m) \bmod 2^{32}$. The seed changes only the initial battery draws. UAV $i$ belongs to battery group $\lfloor (i \bmod 9)/3 \rfloor$, and groups 0, 1 and 2 draw their initial battery from the bands 10\% to 20\%, 45\% to 55\% and 80\% to 90\% of the capacity.

\subsection{Observation}

The 880 inputs are ordered as follows. (1) A 20 by 20 grid of 90~m cells centered on the UAV holds the number of users in each cell and the number of users connected to this UAV in each cell, both divided by 10 (800 numbers). The window includes offsets in $[-900, 900)$~m on each axis, and users outside it are omitted. The two layers are stored one after the other, each row by row along the $y$ axis, and a user at horizontal offset $(\delta_x, \delta_y)$ from the UAV falls into column $\lfloor(\delta_x + 900)/90\rfloor$ and row $\lfloor(\delta_y + 900)/90\rfloor$. (2) Up to 8 other UAVs that have not crashed and lie inside the same window are listed with the same-channel UAVs first. Each of the two groups is sorted by horizontal distance, and ties are resolved by the index of the UAV. Each one is described by its horizontal offsets divided by 1,800, its altitude offset divided by 300, a same-channel flag, a serving flag and a validity flag (48 numbers). Rows without a UAV are zero. (3) The number of users connected to this UAV is divided by 10 (1 number). (4) The position of the UAV is divided by 4,000 (3 numbers). (5) Yang's 19 per-UAV scalars are the grid coordinates divided by 19, a one-hot code of three entries that remains from Yang's three altitude levels (the first entry marks a UAV at a station, the second a UAV in flight, and the third is never set in this simulator), the battery fraction, a one-hot type in the order first type, second type, a one-hot operating state in the order move (which includes flight in charge mode), serve, at a station, crashed, a one-hot state of the UAV's own cell (no station, station full, station with a free slot), the cell of the last station visited divided by 19, and the offset to that station in the same normalized coordinates, which is clipped to $[-1, 1]$. The last station is initially the station from which the UAV starts. (6) The three stations are sorted by distance, with ties in the order of the station index, and each is given by its horizontal offsets and its distance divided by 4,000 (9 numbers).

\subsection{Inactive Reward Terms and Other Remarks}

Yang's exploration and service terms are not active in this simulator, and the reward here has no global coordination term, which his reward contains. His reward of $+1$ for surviving to the final step is never paid, because his code marks a surviving UAV as finished on the final step before it computes that reward. His code also contains a term of $-1$ for a charging violation. The flag that enables it is disabled in the configurations used here, and the handling of arrivals and rejections also prevents the violation. The cap on the SINR remains from an earlier version of the model without a noise term, in which a single transmitter would have had an infinite SINR.

\section{Reproducibility Record}
\label{appx:repro}

This appendix records what the session logs and run directories show beyond the description in Section~\ref{sec:approach}. The LLM calls were made in November 2026. Every session log that we kept shows the model identifier \texttt{gpt-6.1-sol}, version 0.159.2 of the Codex command-line tool, the reasoning effort \texttt{xhigh} and the read-only sandbox. The evaluation scripts of the loops ran on the CPU of a remote host. The session log is the transcript that the command-line tool keeps of each LLM call, including the shell commands that the LLM issued to read files. In the 1,057 session logs of all runs that this paper reports or lists, with 21,649 shell commands, every command reads files or lists paths, apart from 13 \texttt{git} status and branch queries, and none starts an interpreter or runs the environment, the model or the evaluator.

\subsection{What the LLM Could Read and Did Read}

\emph{Single-fleet runs.} The working directory was a checkout of the repository with the simulator source, the task files and the directories of earlier runs, and the task allowed the LLM to read any file. Three specification files that the task names lay outside this checkout, and the LLM read them there. Every single-fleet session also read other files outside the working directory. These included notes that we keep for coding agents, which contain project information, and in some sessions the run configuration of the from-scratch RL baseline. Besides the history file, sessions read the result files of earlier rounds, and one session read the loop log. In every round the LLM read the simulator code that contains the energy and charging logic. The name of the held-out world file appears in file listings of nine sessions, among them the sessions of rounds 4 and 12 of the crash-allowed loop. No session issued a command that printed the content of that file.

\emph{Zero-crash loop.} The text of the zero-crash condition referred to the text of the crash-allowed condition without including it, so the prompts of this loop contained no round number, no path to a history and no stopping rule. In rounds 2, 3 and 6 its sessions read the history, programs or results of the crash-allowed loop, and in round 3 the program and result of the zero-crash one-shot run.

\emph{Game loops.} The working directory was a reduced copy that held the simulator tree, the reference weights, the program that applies a candidate, the specification files and the task text. The held-out world, the top-level test folder, all result folders and every data file that names a fleet score were removed. In our scan of the session logs, the only files that the sessions read outside this directory were instruction files of the command-line tool itself. The specification files inside the directory describe how worlds are generated, so the removal of the held-out world file does not hide all information about layouts. In the 5-UAV runs the task showed the LLM the earlier attempts of the focal UAV, at most 20. In the 30-UAV replacement run it showed the 20 most recent attempts of all focal UAVs. The scores of earlier attempts were measured against the profiles that were in force at the time.

\emph{Self-interested one-shot run.} The working directory was the reduced copy of the game loops, and the task told the LLM to read only files inside it. The session issued 16 shell commands, all on files inside that directory.

\emph{Ablations.} In the two history ablations the working directory was again the whole checkout, so the directories of the earlier runs, including the full history of the crash-allowed loop, were reachable. A text search of the session logs found these paths only in file listings. As in the single-fleet runs, the sessions of the two history ablations also read files outside the working directory, among them our notes for coding agents and, in three sessions, the run configuration of the from-scratch RL baseline. In the two one-shot runs without the reference weight set, the working directory was a copy that contained neither the reference weight set nor the held-out world.

\subsection{Template State Dictionary}

A returned program is applied to a template state dictionary. The single-fleet task describes the template as randomly initialized, and the game tasks describe it as all-zero, but the template of the game loops of Table~\ref{tab:loops} is a checkpoint of the network after four updates of a short training run, and none of its nine actor tensors is all zero. Every returned program that we inspected sets all nine policy tensors to zero before it writes them, so the template does not affect the weights that these programs produce.

\subsection{Evaluation Values}

Table~\ref{tab:fleet} uses evaluations made after the runs with one common configuration. The exception is the guard-off values of the last three RL rows, which are the evaluations made when their training ended. Our record states the same evaluation settings for them, but their command was not kept. For some weight sets the runs' own evaluations at the time gave slightly different values. For the crash-allowed one-shot run, for example, they gave 144.9 instead of 145.1 on the original map and 130.9 instead of 130.2 on the second world. We have not explained these differences. The values in Section~\ref{sec:rounds}, in Fig.~\ref{fig:loop} and in the game loops are the runs' own evaluations.

\subsection{Further Runs}

\emph{Imposed fleet-wide replacement with 5 UAVs.} This run compares the two procedures at the same fleet size as the 5-UAV unilateral run. It used the same configuration, starting profile, worlds and margin, visited UAVs 0 to 4 in order and was set to the same stopping rule. It was stopped manually, with 68 completed attempts and an unfinished attempt 69. Its first candidate was the same program as the first candidate of the 5-UAV unilateral run, apart from the position of two statements, although it came from a separate LLM call. After attempt 1 the fleet had the same five scores as the profile of the 5-UAV unilateral run after attempt 5, with a fleet mean of 90.8. The fleet mean fell to 75.3 in attempt 2 and stayed between 73.2 and 75.8 afterward, below its starting value of 84.7. Of 68 attempts, 66 produced a candidate that was adopted and 2 failed technically, so the adoption test never rejected a candidate. From attempt 2 on, 2.0 to 2.2 of the 5 UAVs per episode crashed on each training world under every adopted profile. From attempt 31 on, the run cycled through the same 5 programs.

Besides the runs of Table~\ref{tab:game} and this run, we ran seven loops with 30 UAVs, static users and imposed fleet-wide replacement. One of them used both training worlds, started from the self-interested weight set of Table~\ref{tab:fleet} and ended with 100 completed attempts and an unfinished attempt 101, without a recorded stopping reason. Six used a single training world and reached their cap of 50 attempts. None had 10 consecutive attempts without an adoption. We also ran two further loops with 5 UAVs and unilateral adoption, which Section~\ref{sec:games} reports as related runs. Finally, we ran four runs with moving users and an adoption margin of 0.03, which this paper does not analyze.

\subsection{Availability of Code and Data}

The ancillary file task-texts.txt reproduces the single-fleet task, the texts of its conditions and the game-specific sections of two game tasks. The tasks of the runs without the reference weight set and of the runs with moving users are not included. This preprint does not provide the simulator, the world files, the candidate programs, the weight files or the run logs, so a reader cannot yet audit the reported runs.

\section{Details of the RL Baselines}
\label{appx:rl}

\emph{Training.} Each policy update collects 250 joint steps in each of 16 parallel copies of the environment, where a joint step is one simulator step in which all 30 UAVs act, so the step counts of Table~\ref{tab:fleet} are 4,000 times the numbers of updates. In all five configurations the training reward of each UAV is its own reward $r_i(t)$ of Eq.~\eqref{eq:reward} and not the fleet-mean score. Where the battery penalty is active, it is subtracted on the steps to which it applies, and the reward of a crash step remains $-6$. At the code commit recorded in each run directory, the parameter file sets a discount factor of 0.99, a generalized-advantage-estimation parameter $\lambda$ of 0.97, learning rates of $10^{-5}$ for the actor and $3 \times 10^{-5}$ for the critic, a clipping range of 0.22 and an entropy coefficient of 0.03, the same values as in Yang's code. We confirmed in the trainer code at these commits that all six values are read. We did not tune these values, and they do not specify the update schedule, the minibatches, the critic inputs or the processing of the advantages.

\emph{Direction mask.} Let $E$ be the battery level before the step and $R$ the reserve of the energy guard. The mask does nothing while $E > 1.5R$. With $\alpha = \operatorname{clip}\big((1.5R - E)/(0.5R), 0, 1\big)$, it restricts the horizontal movement to a sector of half-angle $(1 - \alpha)\pi$ around the direction of a target station, so only that direction remains at $E = R$. The target is the nearest station that has not rejected the UAV and has a free slot. If no such station exists, it is the nearest station that has not rejected the UAV, and if every station has rejected the UAV, it is the nearest station. The factor 1.5 is our choice.

\emph{Battery penalty.} The penalty of 150 applies to a UAV that is in flight before the step and whose battery fraction before the step is below $\tau$, in two cases. In the first case the UAV is farther than 100~m from its nearest station and does not get closer to that station in the step. In the second case charging is selectable before the action, and the UAV does not choose charge. This case includes a UAV that is already in charge mode. The penalty applies at most once per UAV and step, and a crash in the same step still gives the crash reward of $-6$. A penalized step can therefore have a lower reward than the step of a crash when its service component is small. We chose 150 as an order of magnitude. Before these runs, the LLM-written weights had reached fleet-mean scores of 126.4 to 149.2 per UAV and step on the training worlds.

\emph{Warm start.} The starting weights of the two warm-start configurations (Section~\ref{sec:rl}) are taken from our run records. The round-6 weight file of the loop and the copy prepared for these runs have the same SHA-256 hash, but the files that we kept from the run directories do not record which file the runs loaded. During the first 50 updates only the critic is trained. Uchendu et al.~\cite{JSRL} report that a pre-trained actor that is fine-tuned with a randomly initialized critic loses performance, and they use a critic warm-up as a baseline. The number of 50 updates is our choice.

\emph{Evaluation.} The evaluated weight files are named after the final update of each run. For the run from scratch and the warm start with the guard, they have the same SHA-256 hash as the final checkpoints. For the other three runs, the evaluated copies matched the final checkpoints in the first 16 hexadecimal digits of the hash, and the copies were deleted after the evaluation. The evaluator disables the direction mask, and the evaluation configuration that we inspected sets no battery penalty.

\section{Literature Search Record}
\label{appx:search}

The search behind Section~\ref{sec:related} was carried out on November 10, 2026. It used three web-search queries and three keyword searches on the alphaXiv index, which the ancillary file literature-search.txt lists. We then compared the statements that Section~\ref{sec:related} makes about cited sources with the cited passages. This comparison also covered four papers that the index had listed (arXiv 2611.05341, 2611.08070, 2611.09376 and 2611.09382) and arXiv 2512.21024, which the six queries had not returned. We did not record how that paper was found. Of four more papers in the index, we read arXiv 2608.02005 (cited in Section~\ref{sec:related}) and the abstracts of arXiv 2604.00487, 2609.16270 and 2606.05002. According to these abstracts, none of the three has an LLM write network weights or adopt a strategy by a payoff test. We did not search the literature on hyperparameters or architectures proposed by LLMs, and we did not cover program-synthesis systems such as FunSearch, Voyager and Code as Policies.

On November 15, 2026, a scheduled daily query to GPT that the first author had set up listed four papers. One was WeightLoop (arXiv 2611.07319), which Section~\ref{sec:related} already cited. The other three were arXiv 2611.11604, arXiv 2609.19538 and a journal paper in IEEE Transactions on Cognitive Communications and Networking. We read these three, the journal paper in its arXiv version 2603.07456v2, and added them to Section~\ref{sec:related}. We know the schedule and the queries of this daily search only from its own reply and did not verify them.

\section*{Acknowledgment}
The authors used Anthropic's Claude models for text drafting and editing in all sections and appendices of this manuscript, for writing the code that draws Figs.~1 to~6, and for the literature search of Appendix~\ref{appx:search}. They used OpenAI's GPT-6.1 Sol and Claude models to review the text and to check it against the run records. These uses are separate from the LLM of the method in Section~\ref{sec:approach}. The authors reviewed, verified and edited all AI-generated content and take full responsibility for the final publication.

\bibliographystyle{IEEEtran}
\bibliography{ieeeBSTcontrol,references}

\end{document}